\documentclass[10pt,twocolumn,letterpaper]{article}

\usepackage{iccv}              % To produce the CAMERA-READY version
\definecolor{iccvblue}{rgb}{0.21,0.49,0.74}
\usepackage[pagebackref,breaklinks,colorlinks,allcolors=iccvblue]{hyperref}
\usepackage[accsupp]{axessibility}
\usepackage{multirow} % Required for multirows
\usepackage{amssymb}
\usepackage{algorithm}
\usepackage{algorithmic}
\usepackage{graphicx}
\usepackage{amssymb}
\usepackage{transparent}
\usepackage{colortbl}
\usepackage{tikz}
\usepackage{array}
\def\eg{{\it{e.g.}}}

\def\ie{{\it{i.e.}}}

\def\paperID{9462} % *** Enter the Paper ID here
\def\confName{ICCV}
\def\confYear{2025}

\title{From Enhancement to Understanding: Build a Generalized Bridge for Low-light Vision via Semantically Consistent Unsupervised Fine-tuning}

\author{
Sen Wang\textsuperscript{1}\thanks{Equal contribution. This work was done by Sen Wang during an internship at Tencent Youtu Lab.} \quad
Shao Zeng\textsuperscript{2}\footnotemark[1] \quad
Tianjun Gu\textsuperscript{1} \quad
Zhizhong Zhang\textsuperscript{1} \quad
Ruixin Zhang\textsuperscript{2}\footnotemark[2] \quad
Shouhong Ding\textsuperscript{2} \\
Jingyun Zhang\textsuperscript{3} \quad
Jun Wang\textsuperscript{3} \quad
Xin Tan\textsuperscript{1}\thanks{Corresponding author.} \quad
Yuan Xie\textsuperscript{1} \quad
Lizhuang Ma\textsuperscript{1} \\[0.5em]
\begin{large}
\textsuperscript{1}East China Normal University \quad
\textsuperscript{2}Tencent Youtu Lab \quad
\textsuperscript{3}Tencent WeChat Pay Lab 33
\end{large} \\[0em]
}

\begin{document}
\maketitle
\begin{abstract}
Low-level enhancement and high-level visual understanding in low-light vision have traditionally been treated separately. Low-light enhancement improves image quality for downstream tasks, but existing methods rely on physical or geometric priors, limiting generalization. Evaluation mainly focuses on visual quality rather than downstream performance. Low-light visual understanding, constrained by scarce labeled data, primarily uses task-specific domain adaptation, which lacks scalability. To address these challenges, we build a generalized bridge between low-light enhancement and low-light understanding, which we term Generalized Enhancement For Understanding (GEFU). This paradigm improves both generalization and scalability. To address the diverse causes of low-light degradation, we leverage pretrained generative diffusion models to optimize images, achieving zero-shot generalization performance. Building on this, we propose Semantically Consistent Unsupervised Fine-tuning (SCUF). Specifically, to overcome text prompt limitations, we introduce an illumination-aware image prompt to explicitly guide image generation and propose a cycle-attention adapter to maximize its semantic potential. To mitigate semantic degradation in unsupervised training, we propose caption and reflectance consistency to learn high-level semantics and image-level spatial semantics. Extensive experiments demonstrate that our proposed method outperforms current state-of-the-art methods in traditional image quality and GEFU tasks including classification, detection, and semantic segmentation. The code is available at \href{https://github.com/wangsen99/GEFU}{GEFU}.
\end{abstract}    
\section{Introduction}
\label{sec:intro}

\begin{figure}[!t]
\centering
\includegraphics[width=1\linewidth]{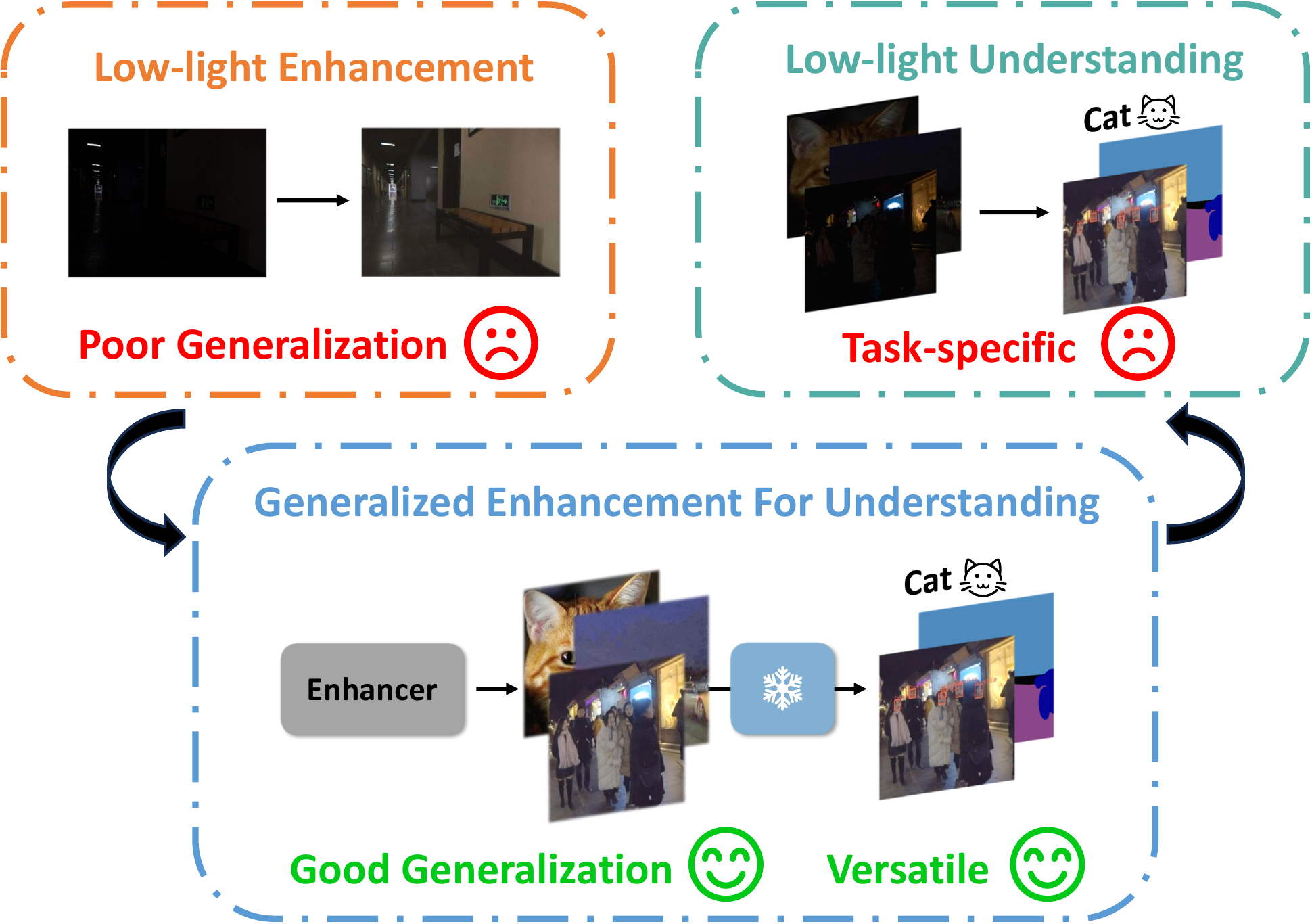}
\caption{We propose ``Generalized Enhancement For Understanding,'' to build a generalized bridge between low-light enhancement and understanding. Face images from~\cite{poor_visibility_benchmark}.}
\label{fig:intro_1}
\vspace{-1em}
\end{figure}

Low-light conditions are unavoidable in the application of visual models, significantly impacting their performance in tasks such as classification~\cite{lengyel2021zero}, segmentation~\cite{Gong2023,wang_fdlnet} and detection~\cite{loh2019getting}.
Research in low-light vision focuses on two main approaches: low-level enhancement and high-level visual understanding. The first approach enhances images based on the physical priors of imaging, but is limited by physical regularization~\cite{guo2020zero, liu2021retinex, ma2022toward} and dataset constraints~\cite{wei2018deepretinexdecompositionlowlight, hai2023r2rnet}, which leads to poor generalization. Additionally, this method focuses on the quality of the image itself, such as the peak signal-to-noise ratio, and evaluation of downstream tasks often requires training visual models from scratch~\cite{liu2021retinex, ma2022toward, cai2023retinexformer}, that is, ``Enhance Then Understand (ETU).'' The second approach involves unsupervised domain adaptation~\cite{lengyel2021zero, luo2023similarity, du2024boosting}, transferring knowledge learned from daytime images to label-poor low-light images for visual tasks. However, this end-to-end learning framework is task-specific and lacks scalability.

Addressing these challenges, we propose a simple yet effective paradigm, ``Generalized Enhancement For Understanding (GEFU),'' which bridges the gap between low-level enhancement and high-level understanding. 
As shown in \cref{fig:intro_1}, this approach includes a robust enhancer that improves images for versatile visual tasks and evaluates the enhanced images directly by testing them on existing pretrained normal light models. 
We compare the evaluation methods of low-light vision in \cref{tab:intro_1}, including low-level image quality and high-level vision tasks. Although some methods try to directly evaluate on downstream tasks~\cite{guo2020zero,jiang2021enlightengan,jiang2024lightendiffusion}, they do not comprehensively consider generalization and effective evaluation methods on downstream tasks.
Compared with ETU, build a framework for GEFU not only extends the limited evaluation methods for low-level enhancement but also improves generalization and scalability through its zero-shot performance across multiple high-level visual tasks.

\begin{table}[!t]
  \centering
  \caption{Comparison of evaluation methods in Low-light Enhancement (LE) methods and Domain Adaptation (DA).}
    \renewcommand{\arraystretch}{1.2}
    \resizebox{\linewidth}{!}{
    \begin{tabular}{r|c|c|ccc|ccc}
    \toprule
    \multicolumn{1}{c|}{\multirow{2}[4]{*}{\textbf{Type}}} & \multirow{2}[4]{*}{\textbf{Method}} & \multirow{2}[4]{*}{\textbf{Image Quatily}} & \multicolumn{3}{c|}{\textbf{Retrain}} & \multicolumn{3}{c}{\textbf{Direct Test}} \\
\cmidrule{4-9}          &       &       & \textbf{Cls} & \textbf{Det} & \textbf{Seg} & \textbf{Cls} & \textbf{Det} & \textbf{Seg} \\
    \midrule
    \multicolumn{1}{c|}{\multirow{15}[2]{*}{LE}} 
    & RetinexNet$^{BMCV'18}$~\cite{wei2018deepretinexdecompositionlowlight} & \checkmark     &       &       &       &       &       &  \\
          & Retinexformer$^{ICCV'23}$~\cite{cai2023retinexformer} & \checkmark     &       & \checkmark     &       &       &       &  \\
          & CIDNet$^{CVPR'25}$~\cite{yan2025hvi} & \checkmark     &       &       &       &       &       &  \\
          & EnlightenGan$^{TIP'21}$~\cite{jiang2021enlightengan} & \checkmark     &       &       &       & \checkmark     &       &  \\
          & Zero-DCE$^{CVPR'20}$~\cite{guo2020zero} & \checkmark     &       &       &       &       & \checkmark     &  \\
          & Zero-DCE++$^{TPAMI'21}$~\cite{li2021learning} & \checkmark     &       &       &       &       & \checkmark     &  \\
          & RUAS$^{CVPR'21}$~\cite{liu2021retinex}  & \checkmark     &       & \checkmark     &       &       &       &  \\
          & SCI$^{CVPR'22}$~\cite{ma2022toward}   & \checkmark     &       & \checkmark     & \checkmark     &       &       &  \\
          & PairLlE$^{CVPR'23}$~\cite{fu2023learning} & \checkmark     &       &       &       &       &       &  \\
          & SADG$^{AAAI'23}$~\cite{zheng2023learning}  &       & \checkmark     & \checkmark     &       &       &       &  \\
          & CLIP-LIT$^{ICCV'23}$~\cite{liang2023iterative} & \checkmark     &       &       &       &       &       &  \\
          & NeRCo$^{ICCV'23}$~\cite{yang2023implicit} & \checkmark     &       &       &       &       &       &  \\
          & QuadPrior$^{CVPR'24}$~\cite{wang2024zero} & \checkmark     &       &       &       &       &       &  \\
          & ZERO-IG$^{CVPR'24}$~\cite{shi2024zero} & \checkmark     &       &       &       &       &       &  \\
          & LightenDiffusion$^{ECCV'24}$~\cite{jiang2024lightendiffusion} & \checkmark     &       &       &       &       & \checkmark     &  \\
    \midrule
    \multicolumn{1}{c|}{\multirow{3}[1]{*}{DA}} & CIConv$^{ICCV'21}$~\cite{lengyel2021zero} &       & \checkmark     & \checkmark     & \checkmark     &       &       &  \\
          & Sim-MinMax$^{ICCV'23}$~\cite{luo2023similarity} &       & \checkmark     & \checkmark     & \checkmark     &       &       &  \\
          & DAI-Net$^{CVPR'24}$~\cite{du2024boosting} &       & \checkmark     & \checkmark     &       &       &       &  \\ 
    \midrule
    \rowcolor{gray!30}
          & Ours  & \checkmark     &       &       &       & \checkmark     & \checkmark     & \checkmark \\
    \bottomrule
    \end{tabular}%
    }
  \label{tab:intro_1}%
  \vspace{-1.8em}
\end{table}%

To learn a robust enhancer, we employ a diffusion generative model like Stable Diffusion (SD)~\cite{rombach2022high} as the image generator, which provides a strong semantic prior to optimize the image. Additionally, we leverage unsupervised learning to fine-tune the generative model, enabling it to achieve zero-shot performance. In SD, text prompt, \eg{,} ``\textit{normal light photo.}" is input into the model to generate semantically consistent results. 
However, this approach leads to two issues. First, the text prompts fail to accurately describe low-light images, reducing them to mere directional guidance with limited semantic value. Second, they lack image-level spatial semantics, which are crucial for capturing details. These limitations in the model framework significantly impact the semantic fidelity of the generated images, particularly in the context of low-light enhancement.

\begin{figure}[!t]
\centering
\includegraphics[width=0.88\linewidth]{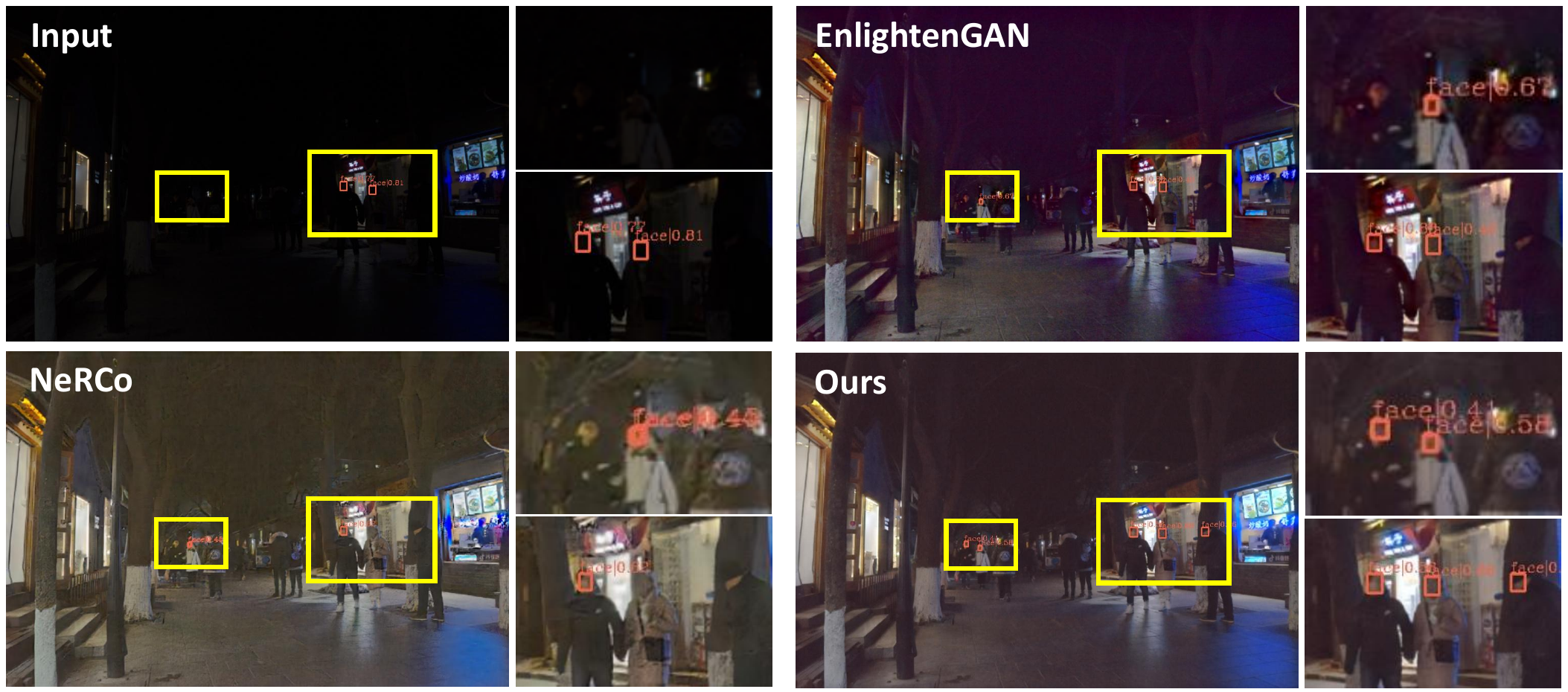}
\caption{Compared with other unsupervised low-light enhancement methods, our approach enhances image semantic fidelity, improving suitability for low-light vision. Face images from~\cite{poor_visibility_benchmark}.}
\label{fig:intro_2}
\vspace{-1em}
\end{figure}

To address these two issues, we propose a Semantically Consistent Unsupervised Fine-tuning method (SCUF) that achieves zero-shot performance and can be seamlessly used for high-level vision tasks.
We first define a cycle generation process including lightening and darkening stages to achieve unsupervised enhancement with SD as the image generator.
For the poor description problem of the text prompt, we introduce an illumination-aware image prompt to explicitly guide image enhancement and a cycle-attention adapter to fully utilize the illumination-aware semantic information of the image prompt.
In addition, we introduce a caption prompt to describe the content of the input image as a supplement to the text prompt and propose caption consistency to ensure high-level abstract semantics consistency in the cycle generation. 
For the problem of losing spatial semantics, we introduce a reflectance map to learn a robust spatial semantic feature representation and propose reflectance consistency to ensure spatial semantics consistency in the cycle generation.
Moreover, caption and reflectance consistency ensure semantic consistency, reducing semantic degradation in the cycle generation.
As shown in \cref{fig:intro_2}, current unsupervised solutions~\cite{jiang2021enlightengan, yang2023implicit} produce artifacts and hinder downstream tasks while our method enhances the semantic fidelity of results. 
% and extensive experiments demonstrate the superiority of our approach in terms of image quality and high-level visual understanding tasks.
% Our method outperforms existing low-light enhancement methods and zero-shot day-night domain adaptation methods in three tasks: night image classification\cite{lengyel2021zero}, dark face detection\cite{poor_visibility_benchmark}, and nighttime semantic segmentation\cite{yu2020bdd100k}.

In summary, our contributions are as follows:
\begin{itemize}
\item We summarize a benchmark for low-light vision evaluation, which we refer to as the ``Generalized Enhancement For Understanding'' task, and introduce a semantically consistent unsupervised fine-tuning framework to improve zero-shot capabilities.
\item We introduce an illumination-aware image prompt to explicitly guide the image generation, and propose a cycle-attention adapter to promote the semantic representation learning.
\item We propose caption consistency and reflectance consistency constitute semantic consistency to learn high-level semantics and image-level spatial semantics in cycle generation, respectively.
\item Extensive experiments have shown that our proposed method is not only superior to current state-of-the-art methods in traditional image quality but also in high-level vision tasks including dark image classification, dark face detection, and nighttime semantic segmentation.
\end{itemize}

\begin{figure*}[!t]
\centering
\includegraphics[width=0.88\linewidth]{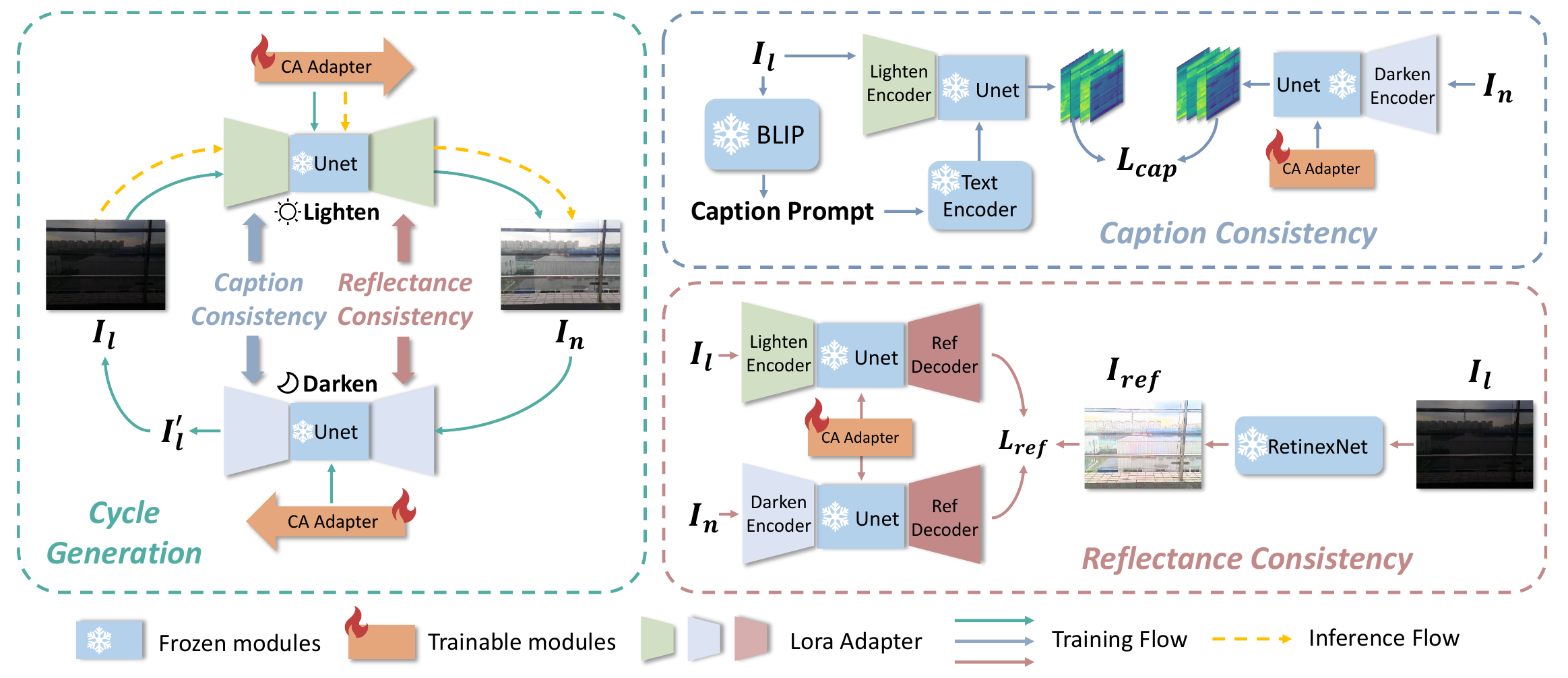}
\caption{The framework of SCUF. The low-light image $\boldsymbol{I_l}$ is input into a \textbf{Cycle Generation} to obtain normal-light image $\boldsymbol{I_n}$ and low-light image $\boldsymbol{I^{'}_l}$ by a cycle loss $\boldsymbol{L_{cycle}}$.  
Reverse and original illumination map from $\boldsymbol{I_l}$ as \textbf{Illumination-Aware Image Prompt} explicitly guide image generation and are fed into a \textbf{Cycle-Attention (CA) Adapter} together with text prompt. 
% \textbf{Caption Consistency (CC)} and \textbf{Reflectance Consistency (RC)} achieve semantic consistency in Cycle Generation. 
\textbf{Caption Consistency} learns a consistent loss $\boldsymbol{L_{cap}}$ of the latent features from the lighten-stage Unet output with caption prompt and the darken-stage output in the Cycle Generation. 
\textbf{Reflectance Consistency} learns a consistency loss $\boldsymbol{L_{ref}}$ of the two stages by additionally introducing a reflectance decoder.}
\label{fig:pl}
\vspace{-1em}
\end{figure*}

\section{Background}
\label{sec:related_work}
% \subsection{Low-light Image Enhancement}
\noindent \textbf{Low-light Image Enhancement} is a classic image processing task. 
Traditional methods are based on handcrafted optimization rules~\cite{bovik2010handbook, park2022histogram, land1977retinex, guo2016lime}. 
Learning-based methods have achieved remarkable results, but many rely on paired low-light and normal-light images~\cite{ zhang2019kindling, zhao2022dsd, Cui_2022_BMVC, wu2023learning, yan2024learnable}, increasing dataset construction costs and risking overfitting.
Unsupervised learning~\cite{wang2024zero, wang2024noise4denoise} mitigate the reliance on paired data, some methods enhance low-light images using illumination~\cite{guo2020zero, li2021learning, shi2024zero} or physical priors~\cite{liu2021retinex, ma2022toward, fu2023learning, wang2024zero}, while others employ generative adversarial networks~\cite{jiang2021enlightengan, yang2023implicit}. Despite their effectiveness within training distributions, these methods struggle with unseen scenes in \cref{fig:intro_2}. 
% To address this, we propose integrating a powerful pre-trained diffusion model into the LIE pipeline.
% Although CLIP-LIT~\cite{liang2023iterative} demonstrates the potential of vision-language models for enhancing backlit images, their application to LIE remains unexplored. 
The ultimate goal of enhancement is for downstream tasks, yet many studies overlook evaluating in this context. While some methods utilize pre-processing step~\cite{cai2023retinexformer} or jointly train models with downstream tasks~\cite{ma2022toward, wang2024unsupervised}, they need to train models from scratch and cannot be applied to multiple tasks.
In contrast, we propose a method that focuses on different visual tasks and achieve zero-shot performance.

\noindent \textbf{Diffusion-based Generative Models} have achieved remarkable success in image generation~\cite{rombach2022high, saharia2022photorealistic, ramesh2022hierarchical, ma2024taming}. 
% Conditional image generation \cite{zhang2023adding} has significantly enhanced the creative potential of the model, while 
Fine-tuning large text-to-image diffusion models has enabled breakthroughs in image editing and related fields~\cite{ruiz2023dreambooth, peng2024portraitbooth, hu2022lora}. However, the extensive iterative steps required by diffusion models hinder real-time applications. The adversarial diffusion distillation~\cite{sauer2024adversarial} addresses this challenge by reducing inference to just 1–4 sampling steps while preserving high-fidelity image generation. Building on this approach, we employ the SD-Turbo to enhance low-light images, significantly improving generation efficiency.

\noindent \textbf{Domain Adaptation} aims to transfer knowledge from a source domain to a target domain~\cite{sakaridis2019guided} and is used for label-scarce scenes.  
% including day-night and zero-shot day-night domain adaptation. 
Day-night domain adaptation leverages both daytime and nighttime images for knowledge transfer~\cite{cui2021multitask, wang2021hla, hoyer2022daformer, hoyer2022hrda, rui_pig}. 
\textbf{Zero-Shot Day-Night Domain Adaptation} tackles nighttime tasks under stricter conditions, without relying on real nighttime images~\cite{lengyel2021zero, luo2023similarity, du2024boosting}. 
For example, DAI-Net~\cite{du2024boosting} incorporates Retinex-based reflectance representation learning to enhance zero-shot low-light object detection. We are inspired by DAI-Net and propose reflectance consistency, however, domain adaptation methods remain inherently task-specific, limiting their general applicability. In contrast, our approach addresses low-light vision through generalized enhancement, enabling seamless integration across various tasks.

\section{Method} \label{sec:method}

% \subsection{Overview} 
%  \label{sec:m_1}
% We propose an unsupervised low-light enhancement framework by utilizing the pre-trained one-step diffusion model SD-Turbo\cite{sauer2023adversarialdiffusiondistillation} as our generator, which is composed of three components: an Encoder, a Unet, and a Decoder.
% based on the image translation model CycleGAN-Turbo\cite{parmar2024one}. The framework 
% Following CycleGAN-Turbo, skip connections are introduced between the encoder and decoder to retain more image details.
% % Then, unlike the general image condition guidance, the conditional image directly is fed into the network to alleviate the conflict between noise and condition.

As shown in \cref{fig:pl}, we propose an unsupervised low-light enhancement framework, which is defined as a cycle generation process.
Given a low-light image $\boldsymbol{I_l}$, we aim to convert it to a normal-light image $\boldsymbol{I_n}$ with our generator based on SD-Turbo~\cite{sauer2023adversarialdiffusiondistillation}, which is composed of three components: an Encoder, a Unet, and a Decoder. 
To this end, we define two objective functions, lightening function $f_{l}(\boldsymbol{I_l},C_l)\colon \boldsymbol{I_l}\to \boldsymbol{I_n}$ and darkening function $f_{d}(\boldsymbol{I_n},C_d)\colon \boldsymbol{I_n}\to \boldsymbol{I_l}$, and corresponding lighten and darken encoders and decoders to learn different latent spaces, where conditional inputs $C_l$ and $C_d$ include the text prompt and the illumination-aware image prompt.
When the input is $\boldsymbol{I_l}$, the cycle generation process can be represented as:
\begin{equation}
\boldsymbol{I^{'}_{l}}=f_{d}(f_{l}(\boldsymbol{I_l},C_l),C_d).
\end{equation}
Then, we use the pre-trained CLIP~\cite{pmlr-clip} to extract image and text features, which will be fed into the proposed Cycle Attention Adapter (CA-Adapter, \cref{sec:m_2}) to enrich latent semantic features and further maintain semantic consistency with multi-query information. 
In addition, given the insufficient semantic detail in text prompts, we propose Caption Consistency (CC, \cref{sec:m_3}), which utilizes the caption of the input image to maintain the high-level semantic consistency of the latent features during cycle generation. 
Moreover, to compensate for the lack of spatial semantic guidance of text and caption prompts, we propose Reflectance Consistency (RC, \cref{sec:m_4}), which learns robust and consistent spatial semantic features. 
During the inference phase, low-light enhancement is achieved by simply inputting the text prompt, the low-light image, and its illumination map into a Lighten Encoder-UNet-Decoder framework.

\begin{figure}[!t]
\centering
\includegraphics[width=0.9\linewidth]{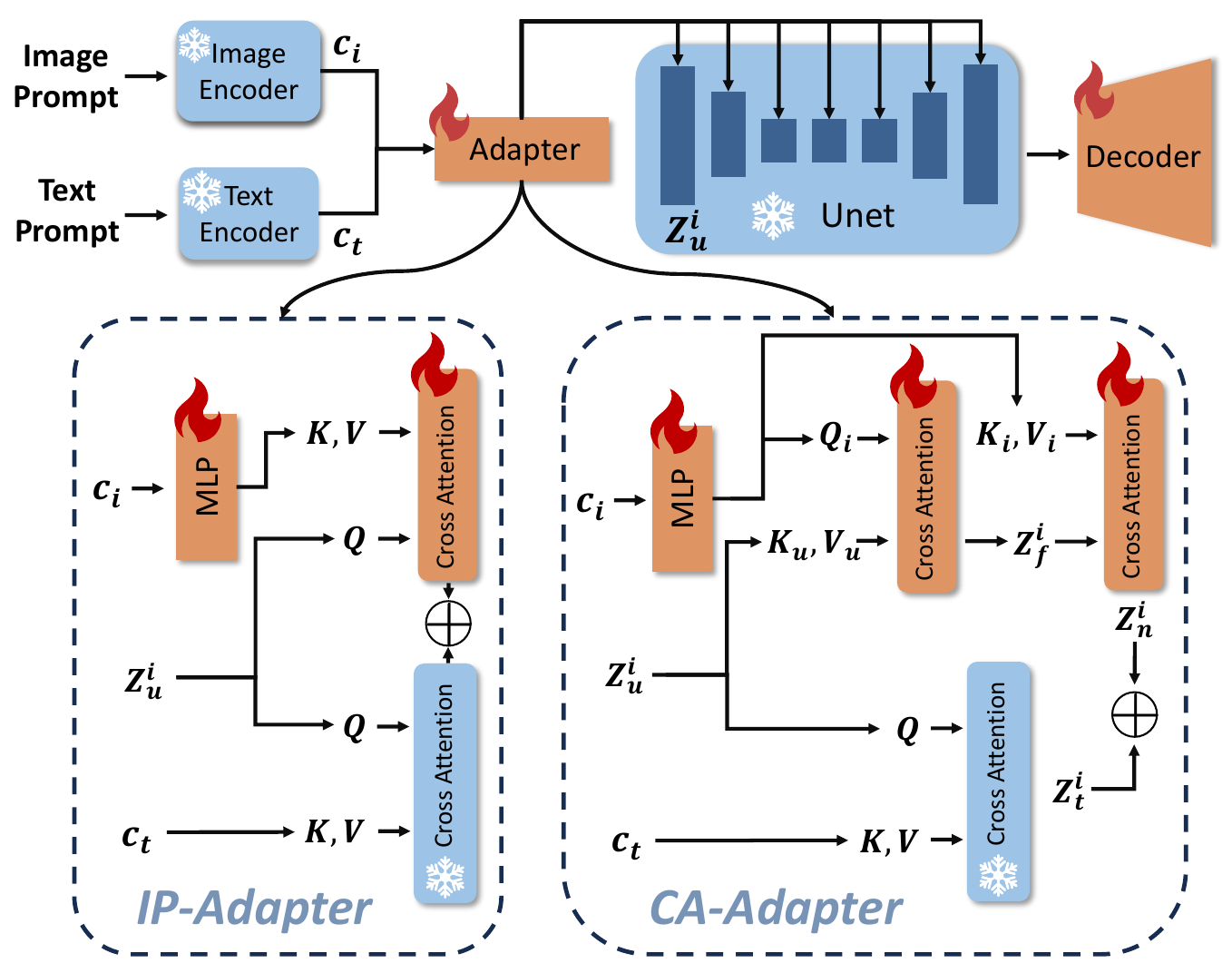}
\caption{Difference between Image Prompt Adapter\cite{ye2023ip} and the proposed Cycle-Attention Adapter.}
\label{fig:ca}
\vspace{-1em}
\end{figure}

\subsection{Cycle-Attention Adapter} 
 \label{sec:m_2}
\textbf{Illumination-aware Image Prompt.} The latent diffusion model uses text as conditional input to generate images~\cite{Rombach_2022_CVPR}. 
However, for low-light images, the textual guidance cannot guarantee image-level semantic alignment. 
We introduce an illumination-aware image prompt to strengthen the semantic guidance of the conditional input by using the illumination map from the original low-light image. 
On the other hand, the illumination-aware image prompt strengthens the attention to the low-light region~\cite{jiang2021enlightengan} and explicitly guides the image generation.
Specifically, when a low-light image $\boldsymbol{I_l}$ is provided as input, it is converted to HSV color space~\cite{gonzales1987digital}. The V channel image $\boldsymbol{I_v}$ is extracted to represent brightness, and the reverse image $\boldsymbol{I_v^r} = 1 - \boldsymbol{I_v}$ is computed. We utilize $\boldsymbol{I_v^r}$ and $\boldsymbol{I_v}$ as illumination-aware image prompts for the lightening and darkening process to generate different illumination images, respectively.

\noindent \textbf{Motivation.} A simple method to insert the image prompt into the pre-trained diffusion model is IP-Adapter~\cite{ye2023ip}. 
As shown in \cref{fig:ca}, IP-Adapter utilizes a decoupled cross attention to integrate text and image prompts respectively, and the query features are all derived from the different scales of latent Unet features $\mathbf{Z_u^i}$ following the Stable Diffusion. 
However, this method of treating text and images equivalently fails to fully leverage the spatial semantic features of images, \ie{,} the abstract descriptions of text features cannot capture the detailed information inherent in images, highlighting the need to fully utilize the rich semantic features of image prompts.

To this end, we propose the Cycle-Attention Adapter to query illumination-aware image prompt features $\boldsymbol{c_i}$ and feed the feature response to itself with another attention layer. 
First, given $\mathbf{Z_u^i}$ from multi-scale Unet features and $\boldsymbol{c_i}$, the output of the first cross-attention layer $\mathbf{Z_f^i}$ can be defined as:
\begin{equation}
    \mathbf{Z_f^i}=\text{Softmax}(\frac{\mathbf{Q_i}\mathbf{K_u}^{\top}}{\sqrt{d}})\mathbf{V_u},
\end{equation}
where $\mathbf{Q_i} = \boldsymbol{c_i}\mathbf{W_{q}^{\prime}}, \mathbf{K_u} = \mathbf{Z_u^i}, \mathbf{V_u} = \mathbf{Z_u}$ are the query, key, and values matrices, and $\mathbf{W_{q}^{\prime}}$ is the learnable linear projection layer. 
Then, we feed the query features and the features responses of $\mathbf{Z_u^i}$ during the first cross-attention back to image prompt features. The final new features $\mathbf{Z_n^i}$ can be defined as:
\begin{equation}
    \mathbf{Z_n^i}=\text{Softmax}(\frac{\mathbf{Z_f^i}\mathbf{K_i}^{\top}}{\sqrt{d}})\mathbf{V_i},
\end{equation}
where $\mathbf{Z_f^i}$ is query, $\mathbf{K_i} = \boldsymbol{c_i}\mathbf{W}_{k}^{\prime}, \mathbf{V_i} = \boldsymbol{c_i}\mathbf{W}_{v}^{\prime}$, and $\mathbf{W}_{k}^{\prime}, \mathbf{W}_{v}^{\prime}$ are the learnable linear projection layers. 
Two cross-attention layers are employed to fully harness the illumination-aware image prompt information.
% By doing this, $c_i$ and $\mathbf{Z}_u$ are cyclically queried in two cross-attention layers, making full use of the prior information of the image prompt.
Finally, the final output of the decoupled cross-attention becomes the sum of the original text cross-attention features $\mathbf{Z_t^i}$ and the new features $\mathbf{Z_n^i}$ output by two cross-attention layers.

\subsection{Caption Consistency} 
 \label{sec:m_3}
\noindent \textbf{Motivation.} We note that the text prompt used in \cref{sec:m_2} is only a description of the target domain, and lacks a detailed depiction of the content in the input image, resulting in limited semantic richness.
To tackle this issue, we introduce a caption prompt as a supplement. However, instead of directly integrating the caption into the text prompt, we keep them separate to prevent semantic degradation during cycle generation, which could hinder the generalization of caption learning. Moreover, this separation enhances inference efficiency.
We copy the encoder of the first stage in the cycle generation process and feed the caption prompt to the cross-attention layer of the initial Unet network to ensure that the output features are consistent with the features of the second stage as shown in \cref{fig:pl}, thereby ensuring the consistency of high-level semantic information during lightening and darkening process.

Taking the input low-light image as an example, we use BLIP~\cite{li2022blip} to generate the caption prompt $C_{cap}$ and input it to the initial Unet $U_{init}$ connected to the lightening encoder $E_l$ and ensure the output is consistent with the Unet $U_{d}$ output connected to the darkening encoder $E_d$, this process can be represented as:
\begin{equation}
    \boldsymbol{\mathcal{L}_{cap, I_l}} = \mathcal{COS}[U_{init}(E_l(\boldsymbol{I_l}),C_{cap}), U_{d}(E_d(\boldsymbol{I_n}),C_d)],
\end{equation}
where $\mathcal{COS}$ represents the cosine similarity, ensuring the consistency of features extracted by the encoder and guiding the CA-Adapter to focus more on high-level semantic information.

\subsection{Reflectance Consistency} 
 \label{sec:m_4}
\noindent \textbf{Motivation.} Caption prompt learns high-level semantic features in the text dimension, and the semantic guidance for image-level enhancement tasks is still limited, so it is necessary to learn the consistency of image-level semantic information. 
We note that some zero-shot domain adaptation frameworks learn Retinex-based~\cite{land1977retinex} reflectance map representations to improve the model performance on high-level vision tasks~\cite{du2024boosting}.
The reflectance map is a counterpart of illumination invariance. Incorporating the reflectance map into the image enhancement process enables the network to learn more robust feature representations.
To this end, we introduce a reflectance map decoder $D_r$ to predict the reflectance map from latent features and propose the reflectance consistency in the cycle generation as shown in \cref{fig:pl} to learn the robust feature extraction of lighten and darken encoders $E_l, E_d$, and CA-Adapters.

Similarly, taking the low-light image as an example, the supervision is divided into two parts, the reflectance map consistency loss $\mathcal{L}_{con}$ and the reconstruction loss $\mathcal{L}_{rec}$, which can be expressed as:
\begin{equation}
\begin{aligned}
\boldsymbol{\mathcal{L}_{ref, I_l}}&=\mathcal{L}_{con}(D_r(Z_l), D_r(Z_d))\\&+\mathcal{L}_{rec}(D_r(Z_d),\boldsymbol{I_{ref}}),
\end{aligned}
\end{equation}
where $Z_l=U_{l}(E_l(\boldsymbol{I_l}),C_l), Z_d=U_{d}(E_d(\boldsymbol{I_n}),C_d)$ represent latent features of Unet outputs in lightening and darkening generation, respectively. 
$I_{ref}$ represents the reflectance map output by a pre-trained RetinexNet\cite{wei2018deepretinexdecompositionlowlight}.  
$\mathcal{L}_{con}$ is the MSE difference and $\mathcal{L}_{rec}$ is the L1 difference. 

\subsection{Unsupervised Training} 
 \label{sec:m_5}
During the training process, we use the LoRA adapter~\cite{hu2021lora} to fine-tune the pre-trained encoders and decoders to construct lightening and darkening latent spaces. Taking the low-light image $I_l$ as an example, the cycle consistency loss can be expressed as:
\begin{equation}
    \boldsymbol{\mathcal{L}_{cycle, I_l}}=\mathcal{L}_{l1}(f_d(f_l(\boldsymbol{I_l},C_l),C_d), \boldsymbol{I_l}),
\end{equation}
where $L_{l1}$ represents the L1 difference. Similar to CycleGAN~\cite{zhu2017unpaired}, we use two adversarial discriminators to classify lightened and darkened results of the model output and real images with a GAN loss $\mathcal{L}_{GAN}$, and for the identity regularization, we also consider its semantic consistency, taking the low-light image as an example, this can be expressed as:
\begin{equation}
    \begin{aligned}
    \boldsymbol{\mathcal{L}_{idt, I_l}}&=\mathcal{L}_{l1}(f_d(\boldsymbol{I_l},C_d),\boldsymbol{I_l})\\& +\mathcal{L}_{cap}(U_{init}(E_l(\boldsymbol{I_l}),C_{cap}), U_{l}(E_d(\boldsymbol{I_l}),C_d))
    \\&+ \mathcal{L}_{rec}(D_r(Z_d),\boldsymbol{I_{ref}}).
    \end{aligned}
\end{equation}
In general, the full objective function incorporates the supervision of unpaired low-light and normal-light images, $I_l$ and $I_n$, respectively. The function is weighted by the parameters $\lambda_{idt}$ and $\lambda_{GAN}$, and is expressed as follows:
\begin{equation}
    \begin{aligned}
\mathcal{L}_{full}&=\boldsymbol{\mathcal{L}_{cycle}} + \boldsymbol{\mathcal{L}_{cap}} + \boldsymbol{\mathcal{L}_{ref}}\\& + 
\lambda_{idt}\mathcal{L}_{idt} + \lambda_{GAN}\mathcal{L}_{GAN}.
    \end{aligned}
\end{equation}

\section{Experiments}

\subsection{Implementation Details}
% We follow CycleGAN-Turbo\cite{parmar2024one} to feed the image to be enhanced into the network directly to alleviate noise and condition conflicts instead of introducing a conditional image encoder. At the same time, 
We utilize skip connections between the encoder and decoder to retain more image details following CycleGAN-Turbo~\cite{parmar2024one} and employ two discriminators built on the CLIP~\cite{pmlr-clip} following the Vision-Aided GAN framework~\cite{kumari2022ensembling}. In the cycle-attention adapter, the illumination-aware image prompt features are uniformly downsampled to be consistent with the Unet features scale.
Our model is trained on the EnlightenGAN dataset~\cite{jiang2021enlightengan}, which includes $\sim$1k unpaired low-light and normal-light images. During training, images are randomly flipped and cropped to a size of 256×256 as input. We use a batch size of 1, following CycleGAN~\cite{zhu2017unpaired}, to ensure training stability. The AdamW optimizer~\cite{loshchilov2018decoupled} is employed with a learning rate of 1e-5. The model is trained for 25k iterations on one RTX 3090 GPU.

\begin{figure*}[!t]
    \centering
    \begin{minipage}{0.3\linewidth}
        \centering
        \begin{subfigure}[b]{0.49\linewidth}
      \includegraphics[width=2.6cm, height=1.2cm]{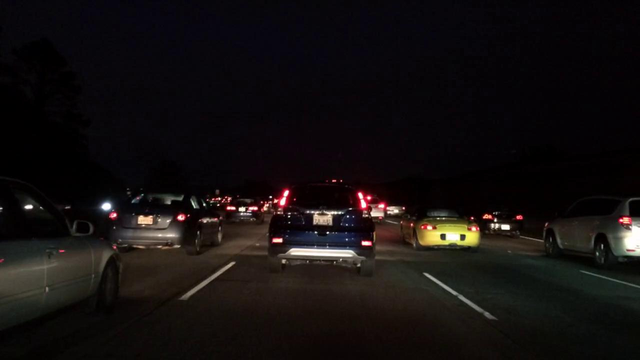}
      \includegraphics[width=2.6cm, height=1.2cm]{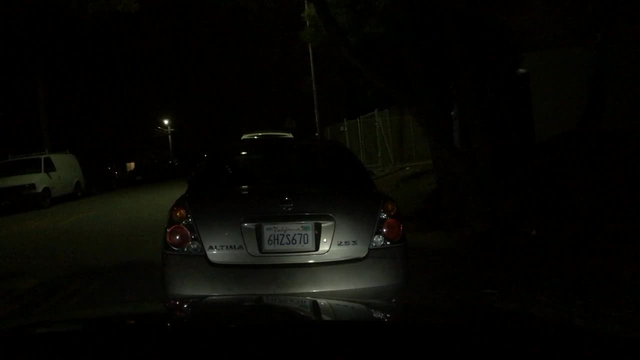}
                \centerline{\footnotesize{(a) Input}}
        \end{subfigure}
        \hfill
        \begin{subfigure}[b]{0.49\linewidth}
      \includegraphics[width=2.6cm, height=1.2cm]{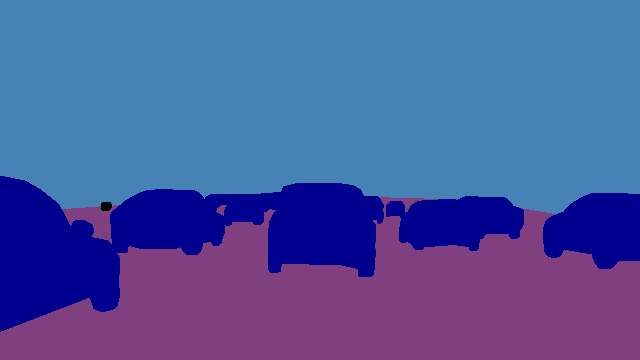}
      \includegraphics[width=2.6cm, height=1.2cm]{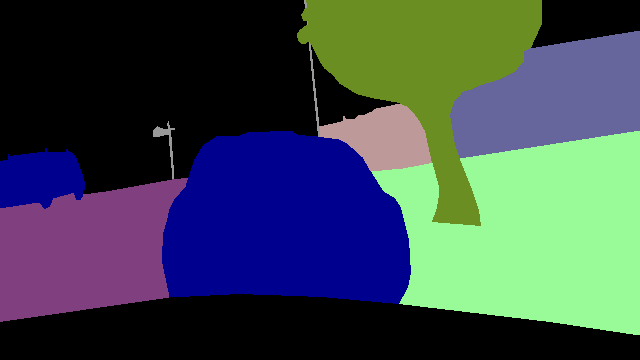}
            \centerline{\footnotesize{(b) GT}}
        \end{subfigure}
    \end{minipage}
    \begin{minipage}{0.3\linewidth}
        \centering
        \begin{subfigure}[b]{0.49\linewidth}
            \includegraphics[width=2.6cm, height=1.2cm]{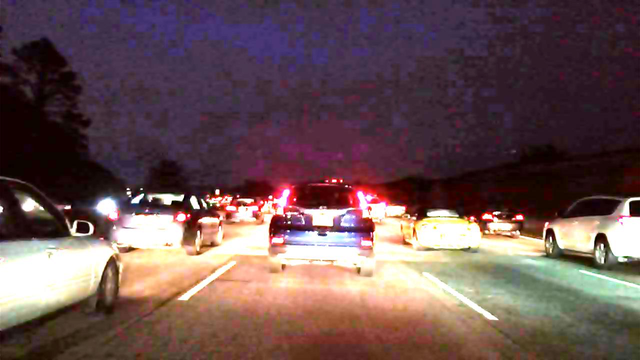}
            \includegraphics[width=2.6cm, height=1.2cm]{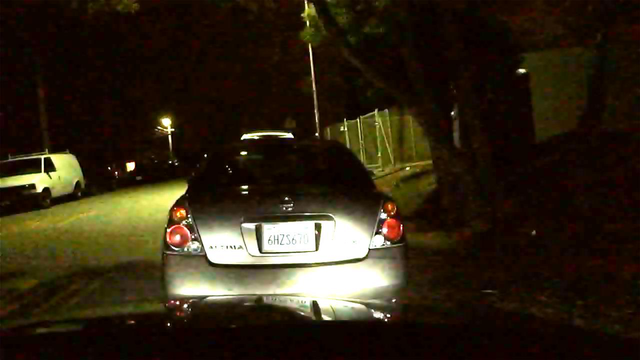}
        \end{subfigure}
        \hfill
        \begin{subfigure}[b]{0.49\linewidth}
            \includegraphics[width=2.6cm, height=1.2cm]{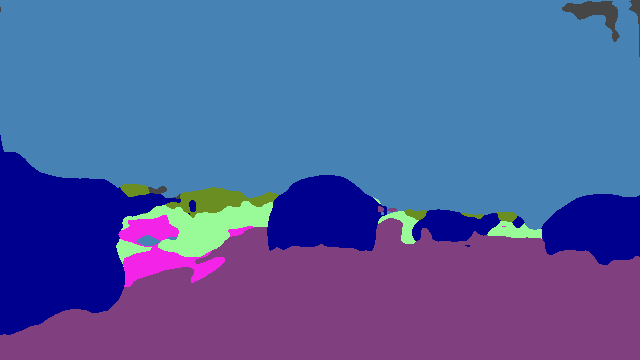}
            \includegraphics[width=2.6cm, height=1.2cm]{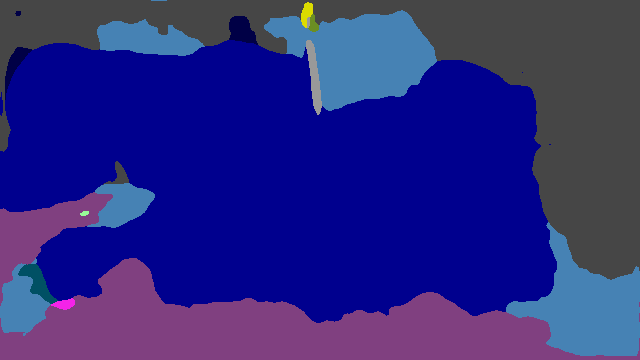}
        \end{subfigure}
        \centerline{\footnotesize{(c) RUAS\cite{liu2021retinex}}}
    \end{minipage}
    \begin{minipage}{0.3\linewidth}
        \centering
        \begin{subfigure}[b]{0.49\linewidth}
            \includegraphics[width=2.6cm, height=1.2cm]{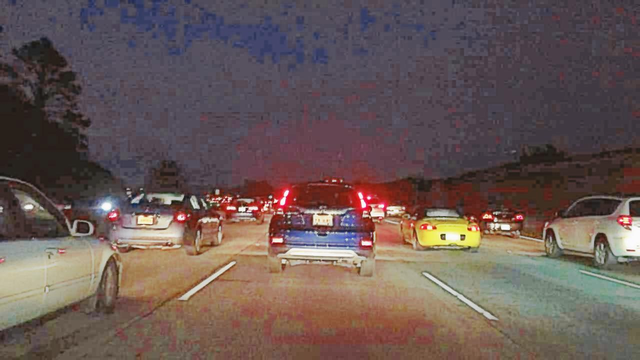}
            \includegraphics[width=2.6cm, height=1.2cm]{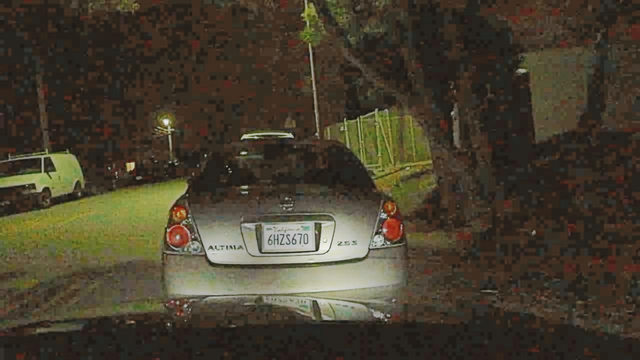}
        \end{subfigure}
        \hfill
        \begin{subfigure}[b]{0.49\linewidth}
            \includegraphics[width=2.6cm, height=1.2cm]{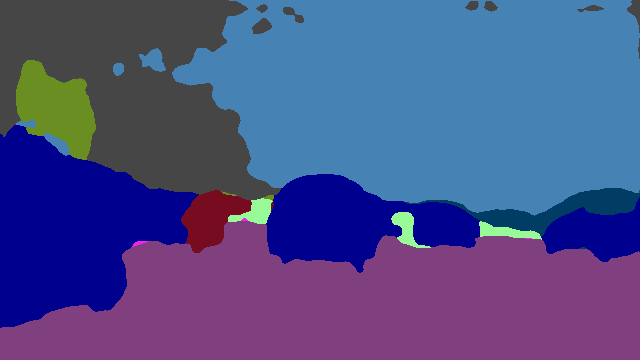}
            \includegraphics[width=2.6cm, height=1.2cm]{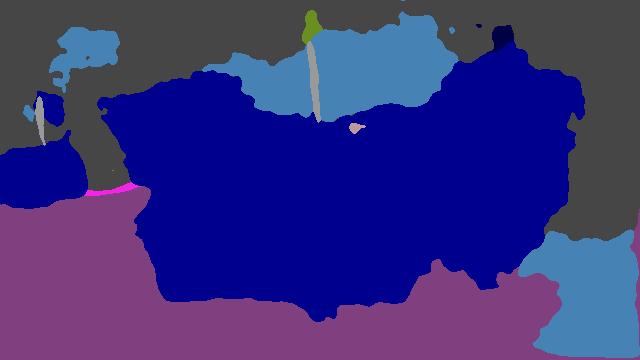}
        \end{subfigure}
        \centerline{\footnotesize{(d) PairLIE\cite{fu2023learning}}}
    \end{minipage}
    \vspace{0.5em}
    
    \begin{minipage}{0.3\linewidth}
        \centering
        \begin{subfigure}[b]{0.49\linewidth}
            \includegraphics[width=2.6cm, height=1.2cm]{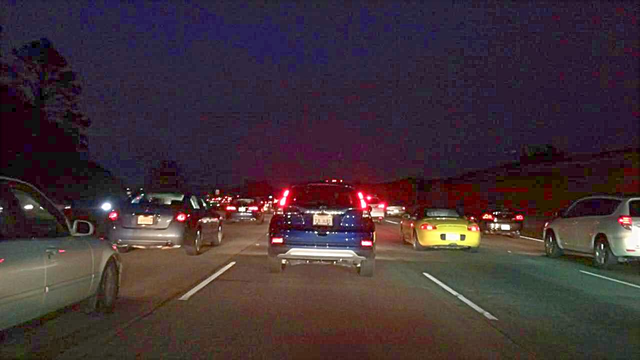}
            \includegraphics[width=2.6cm, height=1.2cm]{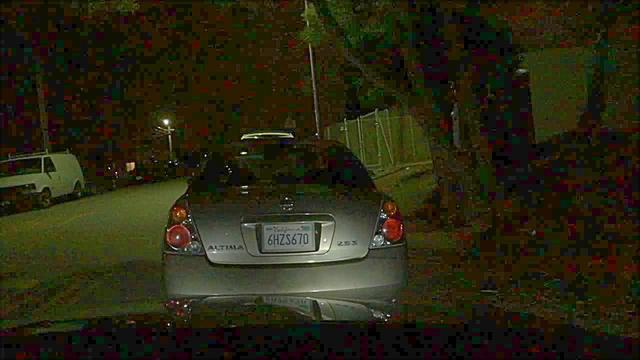}
        \end{subfigure}
        \hfill
        \begin{subfigure}[b]{0.49\linewidth}
            \includegraphics[width=2.6cm, height=1.2cm]{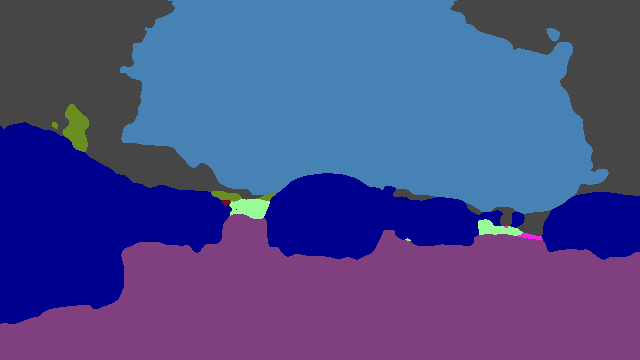}
            \includegraphics[width=2.6cm, height=1.2cm]{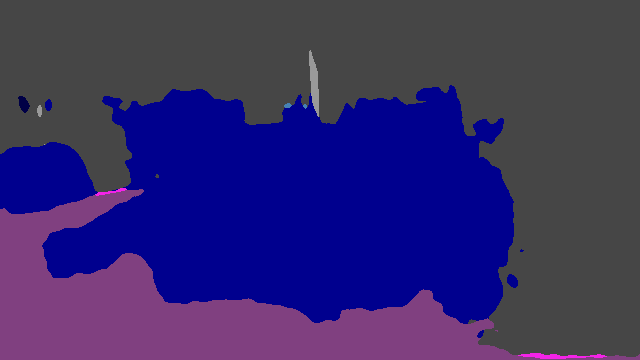}
        \end{subfigure}
        \centerline{\footnotesize{(e) CLIP-LIT\cite{liang2023iterative}}}
    \end{minipage}
    \begin{minipage}{0.3\linewidth}
        \centering
        \begin{subfigure}[b]{0.49\linewidth}
            \includegraphics[width=2.6cm, height=1.2cm]{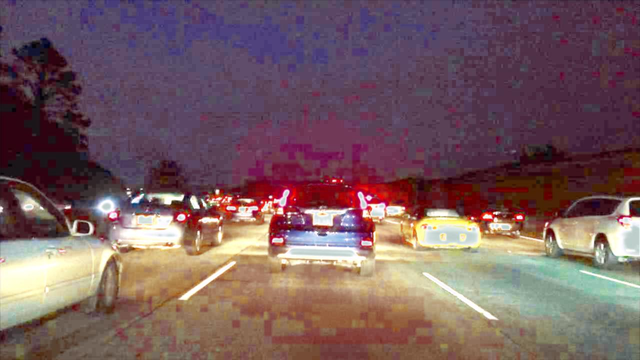}
            \includegraphics[width=2.6cm, height=1.2cm]{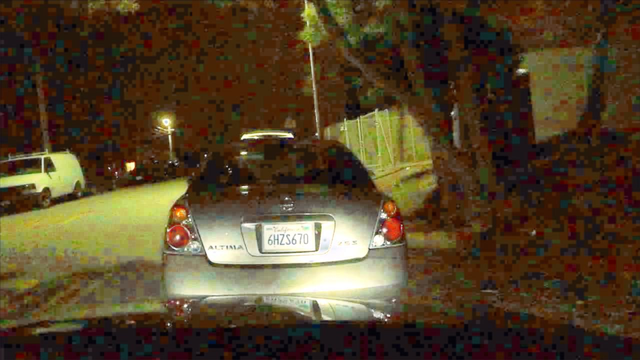}
        \end{subfigure}
        \hfill
        \begin{subfigure}[b]{0.49\linewidth}
            \includegraphics[width=2.6cm, height=1.2cm]{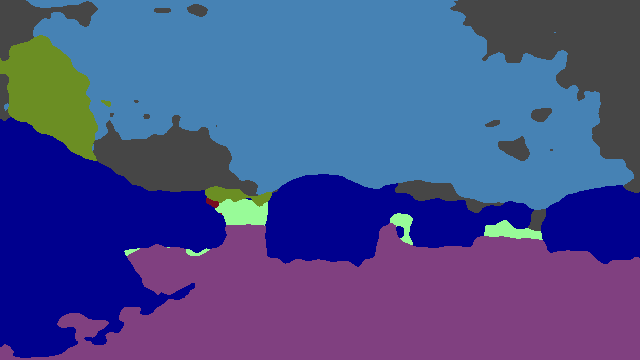}
            \includegraphics[width=2.6cm, height=1.2cm]{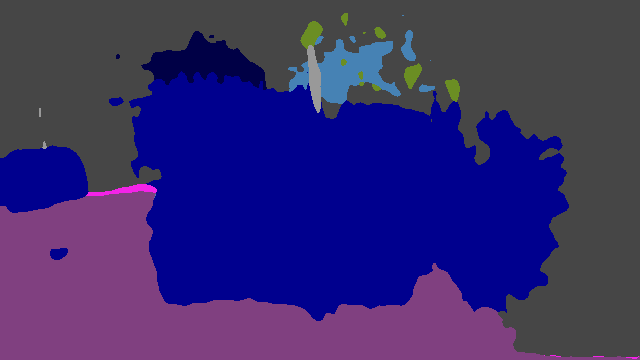}
        \end{subfigure}
        \centerline{\footnotesize{(f) ZERO-IG\cite{shi2024zero}}}
    \end{minipage}
    \begin{minipage}{0.3\linewidth}
        \centering
        \begin{subfigure}[b]{0.49\linewidth}
            \includegraphics[width=2.6cm, height=1.2cm]{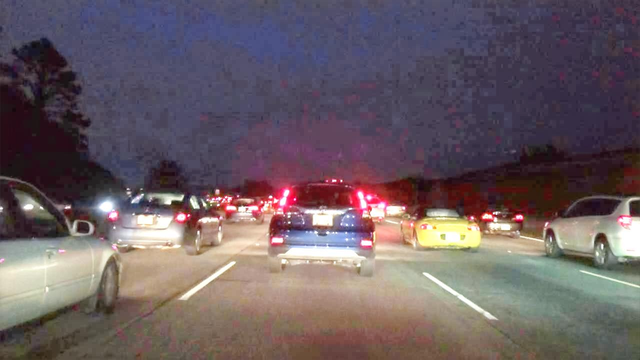}
            \includegraphics[width=2.6cm, height=1.2cm]{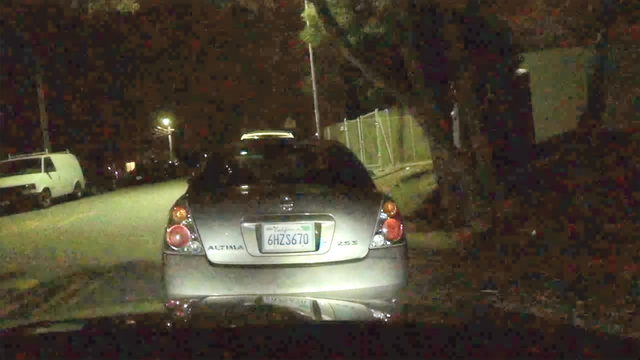}
        \end{subfigure}
        \hfill
        \begin{subfigure}[b]{0.49\linewidth}
            \includegraphics[width=2.6cm, height=1.2cm]{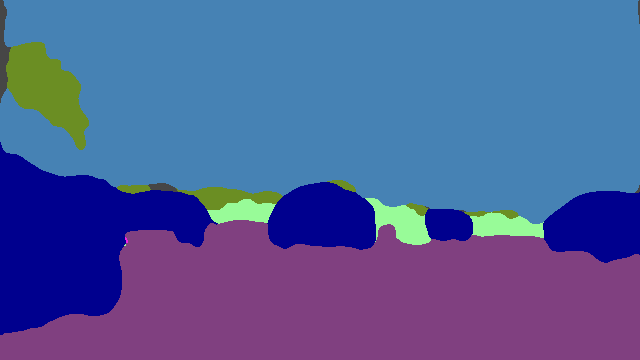}
            \includegraphics[width=2.6cm, height=1.2cm]{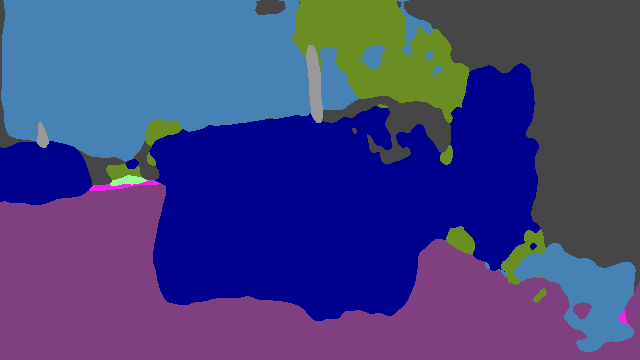}
        \end{subfigure}
        \centerline{\footnotesize{(g) Ours}}
    \end{minipage}

\caption{Qualitative comparison of the proposed method with other state-of-the-art methods on nighttime semantic segmentation.}
\label{fig:ex_2}
\vspace{-1em}
\end{figure*}

\subsection{High-level Vision Comparison}
As shown in \cref{tab:ex_2}, we test three classic visual tasks: classification, detection, and semantic segmentation. 

\noindent \textbf{Classification.} We use the dark image classification dataset CODaN~\cite{lengyel2021zero}, which includes 10,000 daytime image training sets of 10 categories and 2,500 daytime test sets and dark test sets, respectively. 
We follow~\cite{lengyel2021zero} to use ResNet-18~\cite{he2016deep} trained on CODaN daytime data and use the TOP-1 accuracy for evaluation. 

\noindent \textbf{Detection.} We utilize the dark face detection dataset DARK FACE~\cite{poor_visibility_benchmark} includes 10,000 face images captured in real dark light scenes. We follow~\cite{cui2021multitask} to split 600 images as the test set. 
We train the detection model YOLOv3~\cite{farhadi2018yolov3} on WIDER FACE~\cite{yang2016wider} as a baseline and use the mAP for evaluation under the IoU threshold of 0.5.

\noindent \textbf{Semantic Segmentation.} The night images have overexposed and underexposed areas, which are different from low-light vision tasks~\cite{Tan_2021_TIP_NightCity,wang_fdlnet}, so we choose night images in BDD100k~\cite{yu2020bdd100k} that are closer to the low-light domain for testing, named BDD100k-night.
We follow ~\cite{wang_fdlnet} to use 34 night images for testing and use RefineNet~\cite{lin2017refinenet} trained on Cityscapes~\cite{cordts2016cityscapes} as a baseline for evaluation by using mIoU.
\begin{table}[!t]
  \centering
  \caption{Compare with low-light enhancement methods on multiple high-level tasks including classification on CODaN~\cite{lengyel2021zero}, detection on DARK FACE~\cite{poor_visibility_benchmark} and semantic segmentation on BDD100k-night~\cite{wang_fdlnet}. The best results are highlighted in \textbf{bold}. `T', `S', and `U' indicate traditional, supervised, and unsupervised methods, respectively. $\ast$ denotes our re-implementation with the same training data we use.}
  \renewcommand{\arraystretch}{1.1}
    \resizebox{\linewidth}{!}{
  \begin{tabular}{c|l|cc|cc|cc}
    \toprule
    \multirow{3}{*}{\textbf{Type}} & \multirow{2}{*}{Method} 
    & \multicolumn{2}{c|}{CODaN} 
    & \multicolumn{2}{c|}{DARK FACE} 
    & \multicolumn{2}{c}{BDD100k-night} \\
    \cmidrule{3-8}
    & & Top-1(\%) & Time & mAP(\%) & Time & mIoU(\%) & Time \\
    \cmidrule{2-8}
    & Baseline & 53.24 & - & 10.8 & - & 11.4 & - \\
    \midrule
    \multirow{2}[1]{*}{\textbf{T}} 
    & LIME$^{TIP'16}$~\cite{guo2016lime}      & 14.09 & 0.38s & 11.0 & 12.97s & 14.2 & 16.77s \\
    & DUAL$^{CGF'19}$~\cite{zhang2019dual}    & 14.67 & 1.95s & 11.0 & 17min & 14.1 & 21min \\
    \midrule
    \multirow{3}[2]{*}{\textbf{S}} 
    & RetinexNet$^{BMCV'18}$~\cite{wei2018deepretinexdecompositionlowlight}  & 47.48 & - & 13.2 & - & 13.2 & - \\
    & Retinexformer$^{ICCV'23}$~\cite{cai2023retinexformer}  & 52.81 & 0.03s & 16.4 & 0.17s & 15.9 & 0.18s \\
    & CIDNet$^{CVPR'25}$~\cite{yan2025hvi}  & 58.32 & 0.06s & 14.5 & 0.22s & 17.4 & 0.29s \\
    \midrule
    \multirow{15}[2]{*}{\textbf{U}} 
    & EnlightenGan$^{TIP'21}$~\cite{jiang2021enlightengan} & 56.42 & 0.10s & 14.2 & 0.68s & 16.6 & 0.74s \\
    & Zero-DCE$^{CVPR'20}$~\cite{guo2020zero} & 57.76 & 0.02s & 15.9 & \textbf{0.05s} & 16.6 & 0.08s \\
    & Zero-DCE++$^{TPAMI'21}$~\cite{li2021learning} & 59.88 & 0.04s & 15.2 & 0.42s & 17.7 & 0.08s \\
    & RUAS$^{CVPR'21}$~\cite{liu2021retinex}  & 51.60 & 0.06s & 14.0 & 0.48s & 15.2 & 0.30s \\
    & SCI$^{CVPR'22}$~\cite{ma2022toward}   & 58.84 & 0.02s & 14.7 & 0.08s & 18.0 & 0.16s \\
    & PairLIE$^{CVPR'23}$~\cite{fu2023learning} & 52.29 & 0.02s & 16.0 & 1.04s & 16.4 & 0.13s \\
    & SADG$^{AAAI'23}$~\cite{zheng2023learning} & 56.80 & 0.02s & 14.9 & \textbf{0.05s} & 14.8 & 0.08s \\
    & CLIP-LIT$^{ICCV'23}$~\cite{liang2023iterative} & 54.64 & \textbf{0.01s} & 14.1 & 0.39s & 17.3 & \textbf{0.07s} \\
    & NeRCo$^{ICCV'23}$~\cite{yang2023implicit} & 54.15 & 0.10s & 12.4 & 1.28s & 18.1 & 1.54s \\
    & QuadPrior$^{CVPR'24}$~\cite{wang2024zero} & 59.48 & 2.44s & 15.7 & 3.41s & 14.9 & 4.47s \\
    & ZERO-IG-LSRW$^{CVPR'24}$~\cite{shi2024zero} & 47.60 & 0.03s & 15.6 & 0.34s & 14.9 & 0.51s \\
    & ZERO-IG-LOL$^{CVPR'24}$~\cite{shi2024zero} & 53.48 & 0.03s & 15.2 & 0.34s & 14.7 & 0.51s \\
    & LightenDiffusion$^{ECCV'24}$~\cite{jiang2024lightendiffusion} & 57.40 & 0.87s & 16.3 & 1.42s & 16.0 & 1.69s \\
    & LightenDiffusion$^\ast$  & 57.32 & 0.87s & 16.4 & 1.42s & 16.8 & 1.69s \\
    \rowcolor{gray!30}
    & Ours & \textbf{60.92} & 0.18s & \textbf{16.9} & 0.91s & \textbf{20.1} & 0.94s \\
    \bottomrule
  \end{tabular}
  }
  \label{tab:ex_2}
  \vspace{-1.5em}
\end{table}

\noindent \textbf{Comparison Methods.} We comprehensively compare with other low-light enhancement methods including two traditional methods LIME~\cite{guo2016lime}, DUAL~\cite{zhang2019dual}, three supervised methods RetinexNet~\cite{wei2018deepretinexdecompositionlowlight}, Retinexformer~\cite{cai2023retinexformer}, CIDNet~\cite{yan2025hvi} and twelve unsupervised methods EnlightenGan~\cite{jiang2021enlightengan}, Zero-DCE~\cite{guo2020zero}, Zero-DCE++~\cite{li2021learning}, RUAS~\cite{liu2021retinex}, SCI~\cite{ma2022toward}, PairLIE~\cite{fu2023learning}, SADG~\cite{zheng2023learning}, CLIP-LIT~\cite{liang2023iterative}, NeRCo~\cite{yang2023implicit}, QuadPrior~\cite{wang2024zero}, ZERO-IG~\cite{shi2024zero}, LightenDiffusion~\cite{jiang2024lightendiffusion}. 

\begin{table}[!t]
  \centering
  \caption{Compare with zero-shot day-night domain adaptation. The best results are highlighted in \textbf{bold}.}
    \renewcommand{\arraystretch}{1.1}
    \scalebox{0.8}{
        \begin{tabular}{c|c|c}
        \toprule
        \multicolumn{1}{c|}{\multirow{2}{*}{Method}} & DARK FACE & BDD100k-night \\
    \cmidrule{2-3}          & mAP(\%) & mIoU(\%) \\
        \midrule
        Baseline & 16.1  & 11.4 \\
        CIConv~\cite{lengyel2021zero} & 18.4  & 17.6 \\
        Sim-MinMax~\cite{luo2023similarity} & 25.7  & 18.6 \\
        DAI-Net~\cite{du2024boosting} & 28.0    & 18.2 \\
        \midrule
        Ours  & \textbf{29.4}  & \textbf{20.1} \\
        \bottomrule
        \end{tabular}%
      \label{tab:ex_3}%
  }
  \vspace{-1.5em}
\end{table}%

\begin{figure*}[!t]
    \centering
    \begin{minipage}{0.11\linewidth}
      \includegraphics[width=1.9cm, height=1cm]{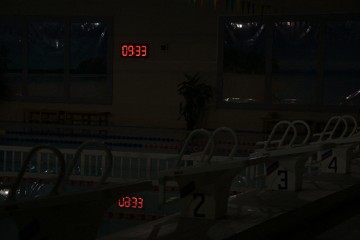}
    \end{minipage}
    % \hfill
    \begin{minipage}{0.11\linewidth}
      \includegraphics[width=1.9cm, height=1cm]{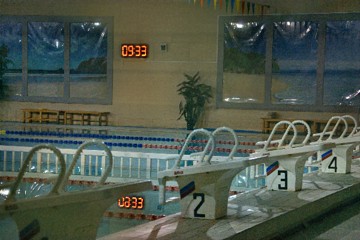}
    \end{minipage}
    % \hfill
    \begin{minipage}{0.11\linewidth}
      \includegraphics[width=1.9cm, height=1cm]{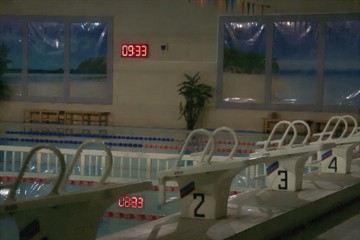}
    \end{minipage}
    \begin{minipage}{0.11\linewidth}
      \includegraphics[width=1.9cm, height=1cm]{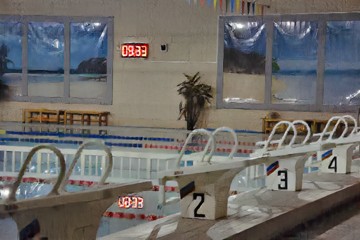}
    \end{minipage}
    % \hfill
    \begin{minipage}{0.11\linewidth}
      \includegraphics[width=1.9cm, height=1cm]{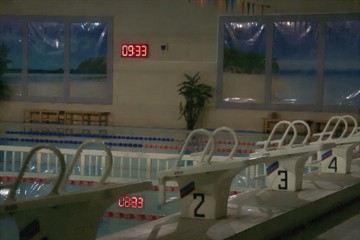}
    \end{minipage}
    \begin{minipage}{0.11\linewidth}
      \includegraphics[width=1.9cm, height=1cm]{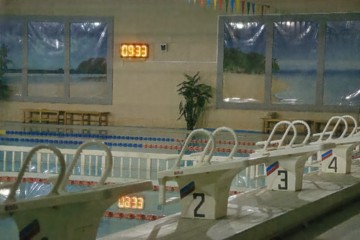}
    \end{minipage}
    % \hfill
    \begin{minipage}{0.11\linewidth}
      \includegraphics[width=1.9cm, height=1cm]{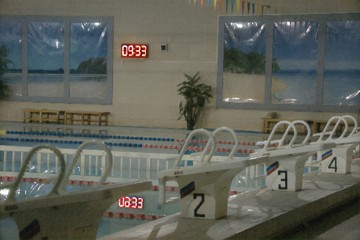}
    \end{minipage}
    \begin{minipage}{0.11\linewidth}
      \includegraphics[width=1.9cm, height=1cm]{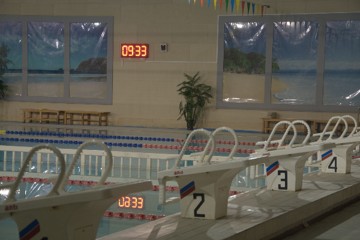}
    \end{minipage}

    \begin{minipage}{0.11\linewidth}
      \includegraphics[width=1.9cm, height=1cm]{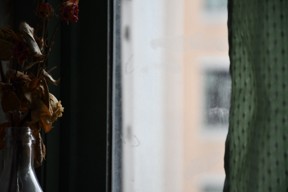}
    \end{minipage}
    % \hfill
    \begin{minipage}{0.11\linewidth}
      \includegraphics[width=1.9cm, height=1cm]{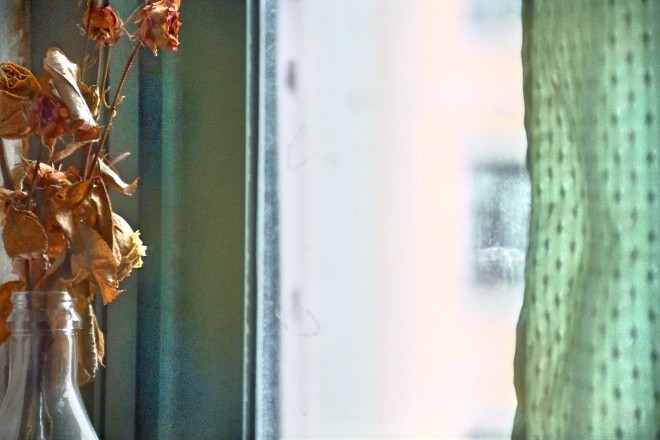}
    \end{minipage}
    % \hfill
    \begin{minipage}{0.11\linewidth}
      \includegraphics[width=1.9cm, height=1cm]{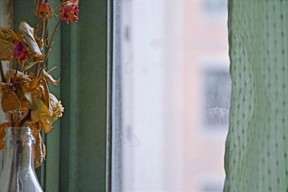}
    \end{minipage}
    \begin{minipage}{0.11\linewidth}
      \includegraphics[width=1.9cm, height=1cm]{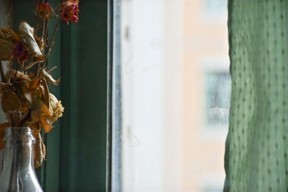}
    \end{minipage}
    % \hfill
    \begin{minipage}{0.11\linewidth}
      \includegraphics[width=1.9cm, height=1cm]{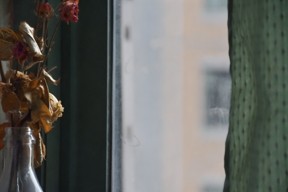}
    \end{minipage}
    \begin{minipage}{0.11\linewidth}
      \includegraphics[width=1.9cm, height=1cm]{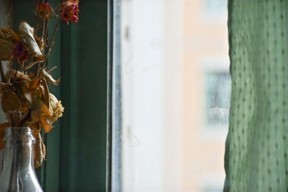}
    \end{minipage}
    % \hfill
    \begin{minipage}{0.11\linewidth}
      \includegraphics[width=1.9cm, height=1cm]{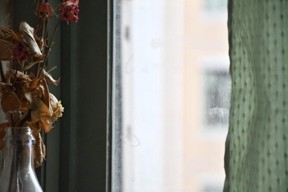}
    \end{minipage}
    \begin{minipage}{0.11\linewidth}
      \includegraphics[width=1.9cm, height=1cm]{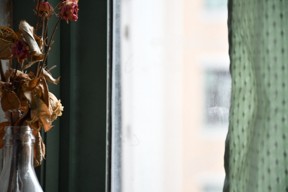}
    \end{minipage}

    \begin{minipage}{0.11\linewidth}
      \includegraphics[width=1.9cm, height=1cm]{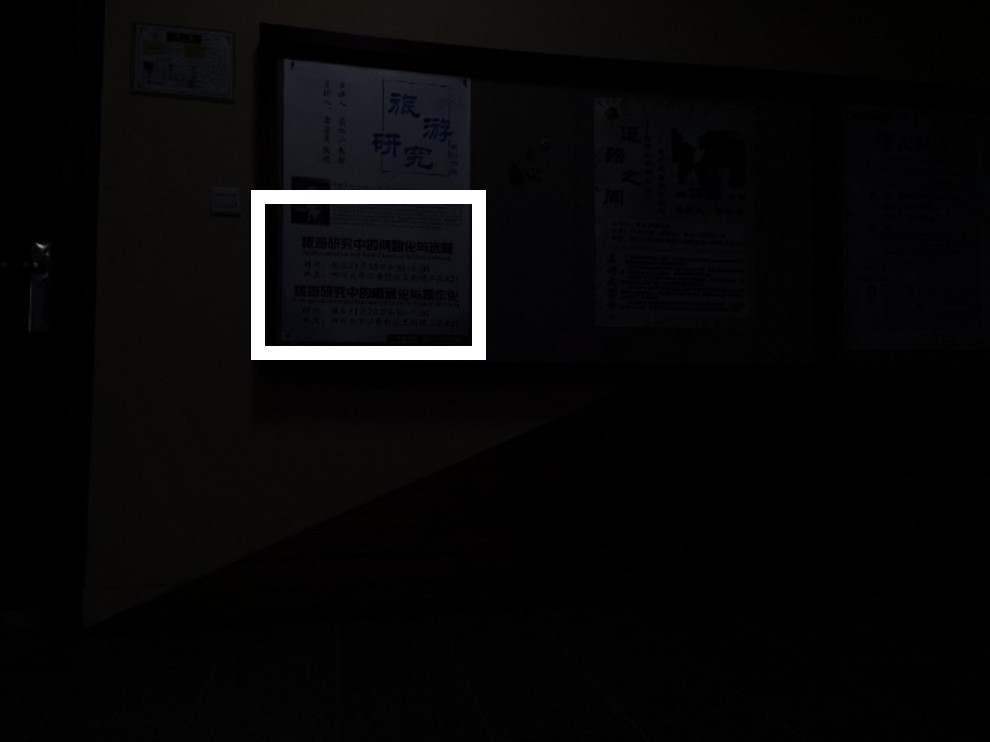}
    \end{minipage}
    % \hfill
    \begin{minipage}{0.11\linewidth}
      \includegraphics[width=1.9cm, height=1cm]{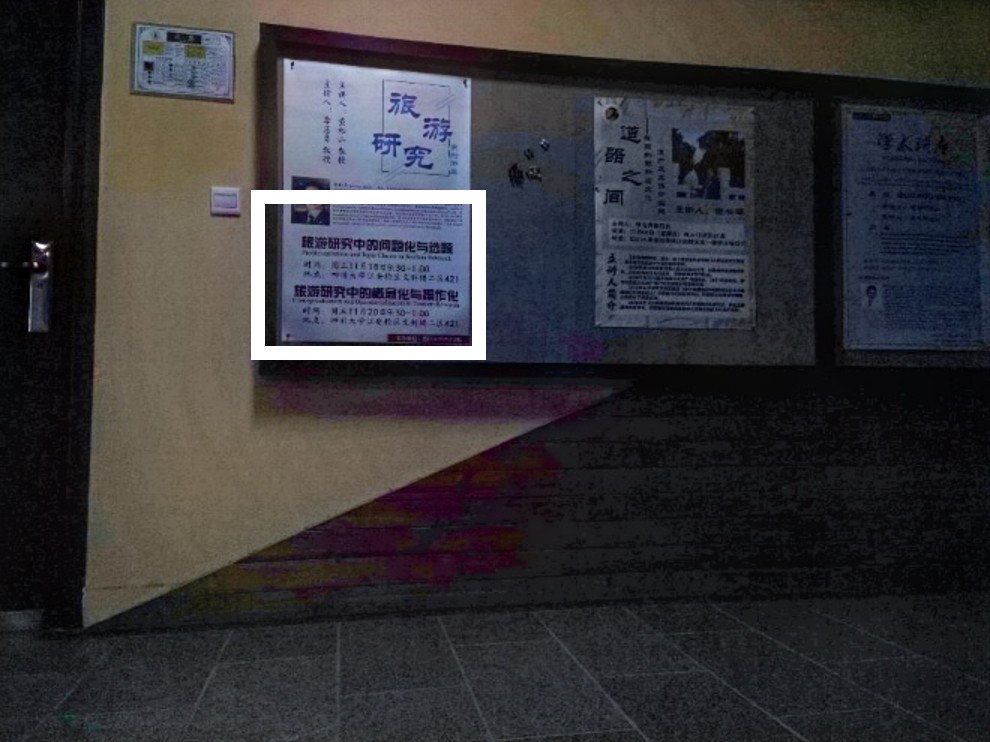}
    \end{minipage}
    % \hfill
    \begin{minipage}{0.11\linewidth}
      \includegraphics[width=1.9cm, height=1cm]{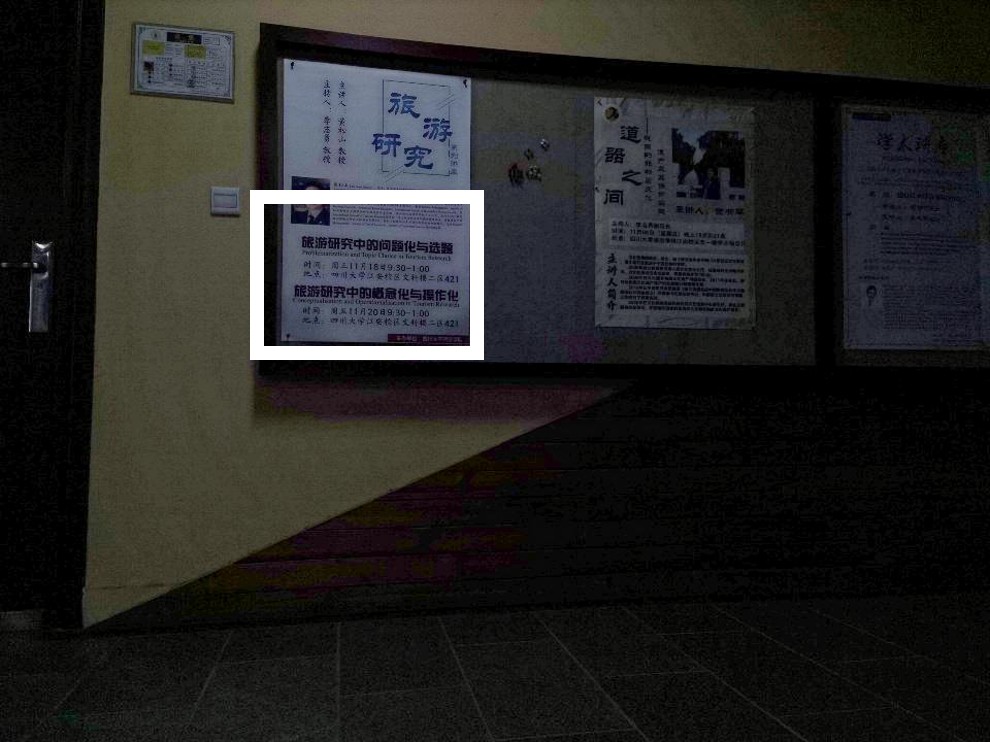}
    \end{minipage}
    \begin{minipage}{0.11\linewidth}
      \includegraphics[width=1.9cm, height=1cm]{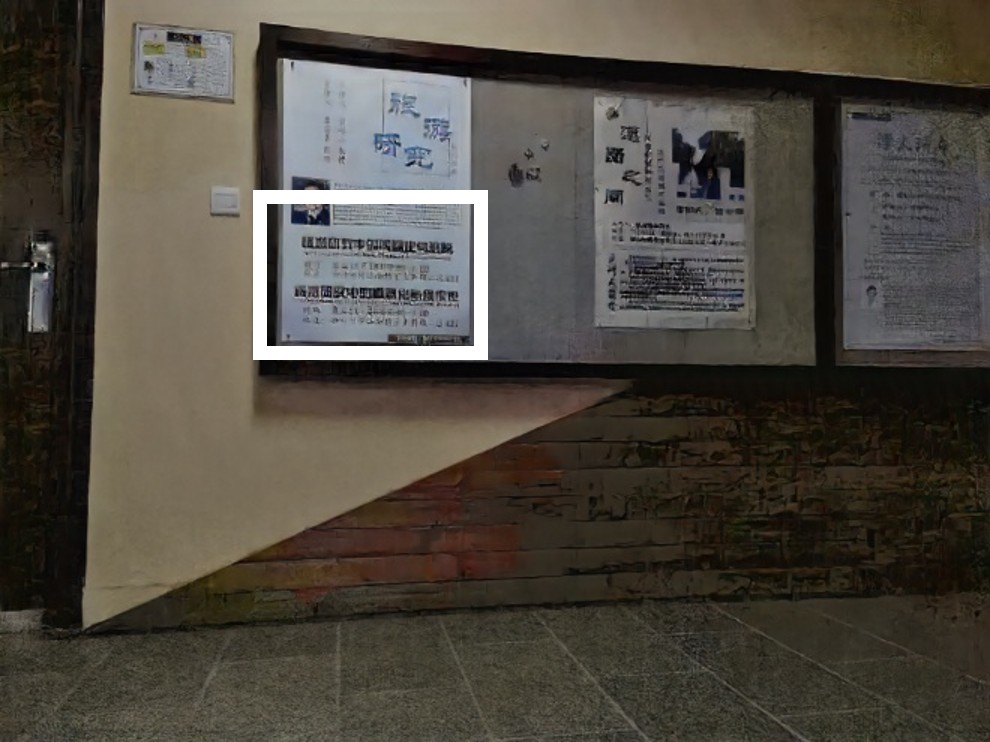}
    \end{minipage}
    % \hfill
    \begin{minipage}{0.11\linewidth}
      \includegraphics[width=1.9cm, height=1cm]{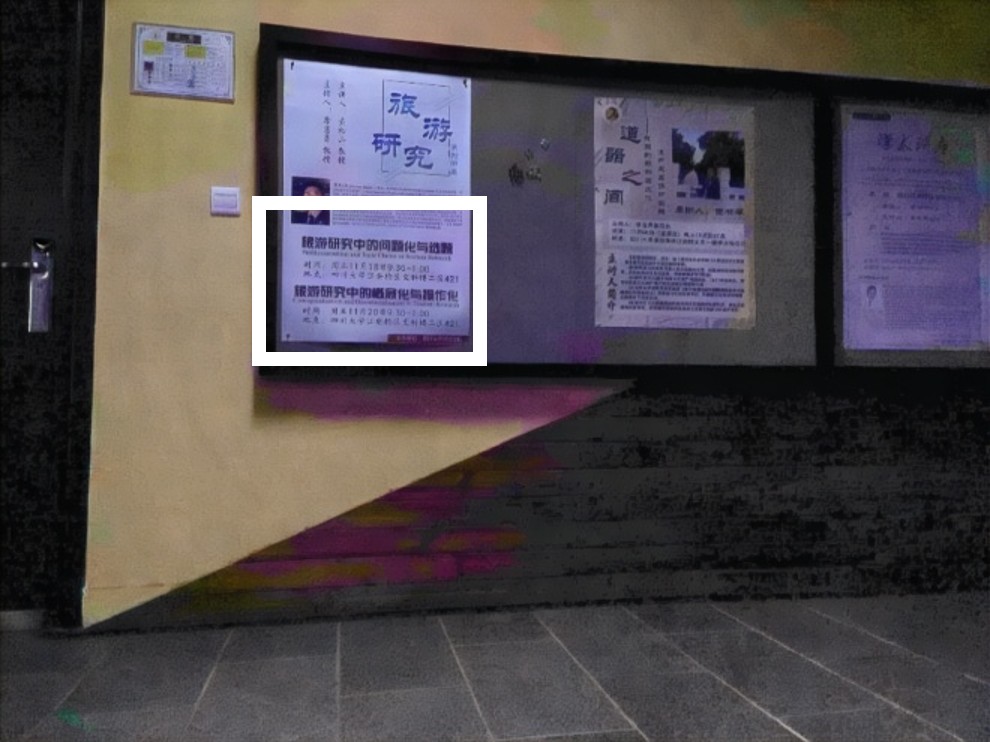}
    \end{minipage}
    \begin{minipage}{0.11\linewidth}
      \includegraphics[width=1.9cm, height=1cm]{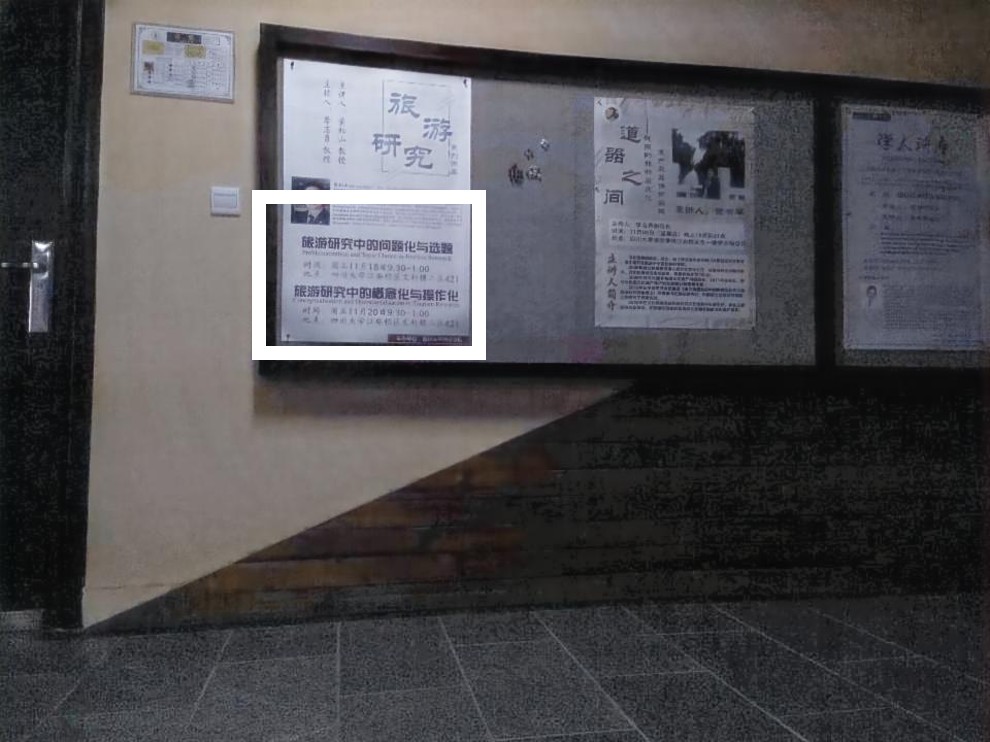}
    \end{minipage}
    % \hfill
    \begin{minipage}{0.11\linewidth}
      \includegraphics[width=1.9cm, height=1cm]{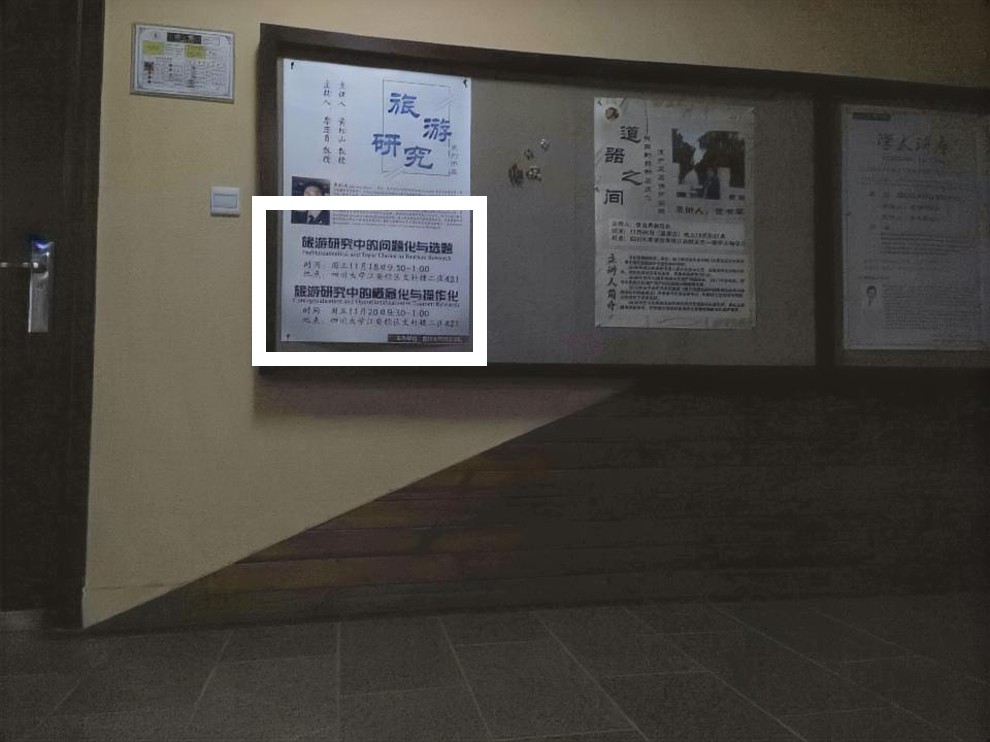}
    \end{minipage}
    \begin{minipage}{0.11\linewidth}
      \includegraphics[width=1.9cm, height=1cm]{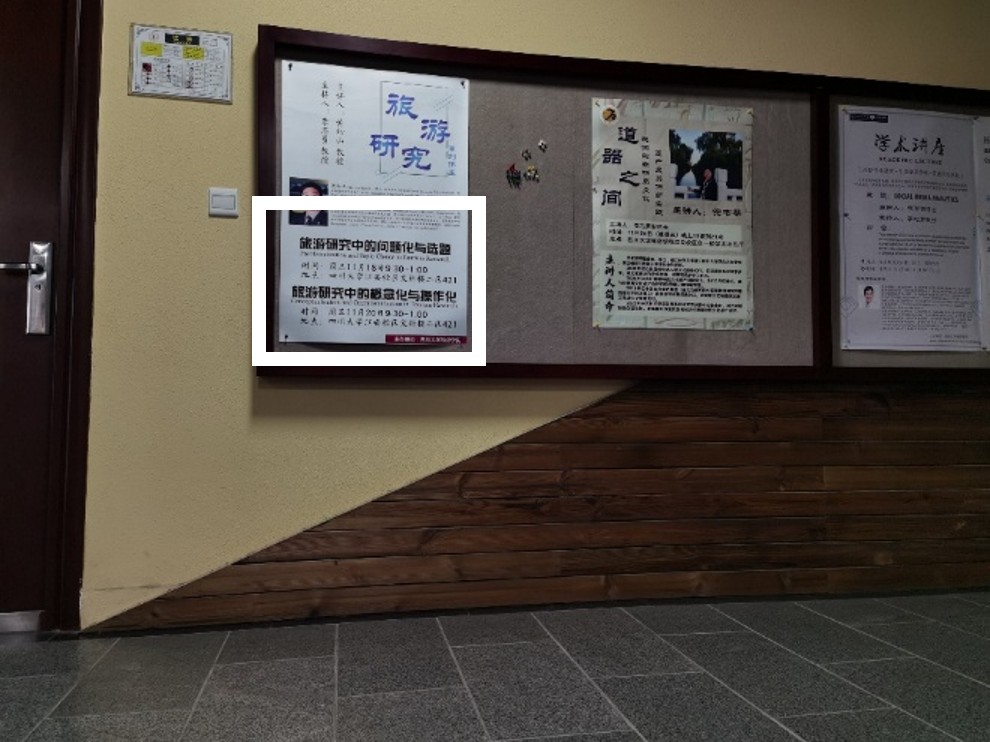}
    \end{minipage}
    
    \begin{minipage}{0.11\linewidth}
      \includegraphics[width=1.9cm, height=1cm]{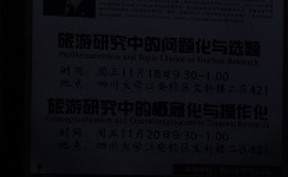}
      \centerline{\footnotesize{Input}}
    \end{minipage}
    % \hfill
    \begin{minipage}{0.11\linewidth}
      \includegraphics[width=1.9cm, height=1cm]{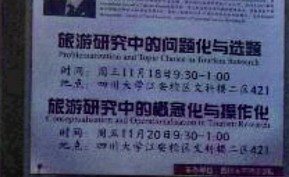}
      \centerline{\footnotesize{EnlightenGan\cite{jiang2021enlightengan}}}
    \end{minipage}
    % \hfill
    \begin{minipage}{0.11\linewidth}
      \includegraphics[width=1.9cm, height=1cm]{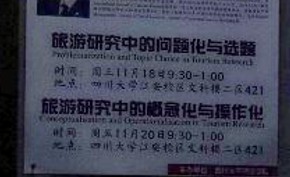}
      \centerline{\footnotesize{Zero-DCE\cite{guo2020zero}}}
    \end{minipage}
    \begin{minipage}{0.11\linewidth}
      \includegraphics[width=1.9cm, height=1cm]{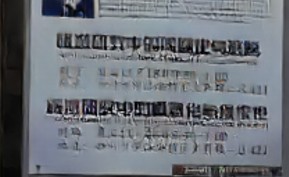}
      \centerline{\footnotesize{NeRCo\cite{yang2023implicit}}}
    \end{minipage}
    % \hfill
    \begin{minipage}{0.11\linewidth}
      \includegraphics[width=1.9cm, height=1cm]{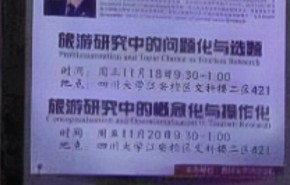}
    \centerline{\footnotesize{QuadPrior\cite{wang2024zero}}}
    \end{minipage}
    \begin{minipage}{0.11\linewidth}
      \includegraphics[width=1.9cm, height=1cm]{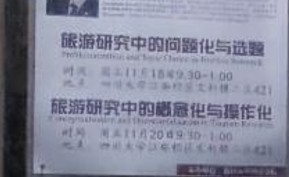}
    \centerline{\footnotesize{LightenDiffusion\cite{jiang2024lightendiffusion}}}
    \end{minipage}
    % \hfill
    \begin{minipage}{0.11\linewidth}
      \includegraphics[width=1.9cm, height=1cm]{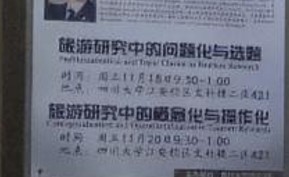}
    \centerline{\footnotesize{Ours}}
    \end{minipage}
    \begin{minipage}{0.11\linewidth}
      \includegraphics[width=1.9cm, height=1cm]{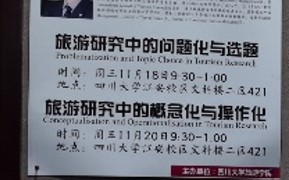}
    \centerline{\footnotesize{GT}}
    \end{minipage}

\caption{Visual quality comparison between the proposed method and other state-of-the-art methods on the LOL\cite{yang2021sparse} and LSRW\cite{hai2023r2rnet}.}
\label{fig:ex_1}
\vspace{-1em}
\end{figure*}

\noindent \textbf{Comparison with Low-light Enhancement Methods.} 
All methods enhance images first and then we utilize the corresponding pre-trained normal light models to test directly on different tasks. 
More complete results are shown in the \textit{supplementary material}.
% We notice that the proposed method trained on EnlightenGAN data had better generalization performance in visual quality comparison than the model trained on LSRW, so we use the former for testing. 
As shown in \cref{tab:ex_2}, most methods have improved the classification, detection, and segmentation performance of the model to varying degrees, and our method achieved the best performance in all three tasks. 
\cref{fig:ex_2} shows that our method enhances the image with the best semantic quality. 
We also compare inference speed of different methods, and our method is not the fastest, but its generalization performance is significantly better than lightweight models such as Zero-DCE and SCI.
% Even if some regions have labeling errors, our method can correctly segment them compared to other methods, \eg{,} the tree on the left side of the first image and the sky in the second image.
% For example, NeRCo\cite{yang2023implicit} performs well in LSRW, but its performance in dark face detection with high local detail requirements is not good. Similarly, LightenDiffusion\cite{jiang2024lightendiffusion} performs well in LOL, but its performance improvement in classification and segmentation tasks is limited.

\noindent \textbf{Comparison with Zero-shot Day-night Domain Adaptation Methods.} To further demonstrate the superiority of our method, we also compared it with state-of-the-art zero-shot day-night domain adaptation methods, including CIConv~\cite{lengyel2021zero}, Sim-MinMax~\cite{luo2023similarity} and DAI-Net~\cite{du2024boosting}. 
For a fair comparison, we follow the DAI-Net setup using the face detection model DSFD~\cite{li2019dsfd} to evaluate DARK FACE, while semantic segmentation is performed according to our setup. 
As shown in \cref{tab:ex_3}, our method achieves the best performance in both tasks, although their methods require training from scratch for different tasks, while we directly enhance and test low-light images in high-level vision.

\begin{table}[!t]
  \centering
  \caption{Image visual quality comparison on LOL~\cite{yang2021sparse} and LSRW~\cite{hai2023r2rnet} datasets. We highlight the top-ranking scores in \textcolor{red}{red} and the second in \textcolor{blue}{blue}. $\ast$ denotes our re-implementation with the same training data we use.}
    \renewcommand{\arraystretch}{1.1}
    \resizebox{\linewidth}{!}{
    \begin{tabular}{c|l|ccc|ccc}
    \toprule
    \multirow{2}{*}{\textbf{Type}} & \multirow{2}{*}{\textbf{Method}} & \multicolumn{3}{c|}{LSRW} & \multicolumn{3}{c}{LOL} \\
\cmidrule{3-8}          &       & PSNR $\uparrow$ & SSIM $\uparrow$ & LPIPS $\downarrow$ & PSNR $\uparrow$ & SSIM $\uparrow$ & LPIPS $\downarrow$ \\
    \midrule
    \multirow{2}[2]{*}{\textbf{T}} & LIME$^{TIP'16}$~\cite{guo2016lime}    & 14.88  & 0.3487  & 0.4030      & 16.90  & 0.4917   & 0.4022 \\
          & DUAL$^{CGF'19}$~\cite{zhang2019dual}  & 13.76  & 0.3532  & 0.4150  & 16.76  & 0.4911  & 0.4060 \\
    \midrule
    \multirow{2}[2]{*}{\textbf{S}} & RetinexNet$^{BMCV'18}$~\cite{wei2018deepretinexdecompositionlowlight}  & 15.59   & 0.4176      & 0.3998  & 17.68   & 0.6477   & 0.4433 \\
          & Retinexformer$^{ICCV'23}$~\cite{cai2023retinexformer}   & 17.19   & 0.5093   & 0.3314    & 22.79     & 0.8397  & 0.1707 
          \\
          & CIDNet$^{CVPR'25}~\cite{yan2025hvi}$ & 18.00 & 0.5198 & 0.2962 & 20.68 & 0.8411 & 0.1156 \\
    \midrule
    \multirow{16}[2]{*}{\textbf{U}} & EnlightenGan$^{TIP'21}$~\cite{jiang2021enlightengan} & 17.59 & 0.4867 & 0.3117 & 18.68 & 0.6728 & 0.3013 \\
          & Zero-DCE$^{CVPR'20}$~\cite{guo2020zero} & 15.86 & 0.4536 & 0.3176 & 18.06 & 0.5736 & 0.3125 \\
          & Zero-DCE++$^{TPAMI'21}$~\cite{li2021learning} & 16.21 & 0.4571 & 0.3266 & 17.37 & 0.4373 & 0.3118 \\
          & RUAS$^{CVPR'21}$~\cite{liu2021retinex}  & 14.33 & 0.4841 & 0.4800  & 15.33 & 0.4876 & 0.3097 \\
          & SCI$^{CVPR'22}$~\cite{ma2022toward}   & 15.24 & 0.4240 & 0.3218 & 17.30  & 0.5335 & 0.3079 \\
          & PairLIE$^{CVPR'23}$~\cite{fu2023learning} & 17.60  & 0.5118 & 0.3290 & 19.88 & 0.7777 & 0.2341 \\
          & SADG$^{AAAI'23}$~\cite{zheng2023learning}  & 16.32 & 0.4564 & 0.3471 & 16.93 & 0.5372 & 0.3513 \\
          & CLIP-LIT$^{ICCV'23}$~\cite{liang2023iterative} & 13.47 & 0.4089 & 0.3572 & 15.18 & 0.5290 & 0.3689 \\
          & NeRCo$^{ICCV'23}$~\cite{yang2023implicit} & \textcolor{red}{19.46} & \textcolor{blue}{0.5506} & 0.3052 & 19.66 & 0.7172 & 0.2705 \\
          & QuadPrior$^{CVPR'24}$~\cite{wang2024zero} & 16.90  & 0.5429 & 0.3459 & \textcolor{blue}{20.30}  & \textcolor{blue}{0.7909} & \textcolor{red}{0.1858} \\
          & ZERO-IG-LSRW$^{CVPR'24}$~\cite{shi2024zero} & 18.21 & \textcolor{red}{0.5665} & 0.4946 & 18.65 & 0.4819 & 0.3819 \\
          & ZERO-IG-LOL$^{CVPR'24}$~\cite{shi2024zero} & 16.44 & 0.5033 & 0.3744 & 18.13 & 0.7455 & 0.2478 \\
          & LightenDiffusion$^{ECCV'24}$~\cite{jiang2024lightendiffusion} & 18.42 & 0.5334 & 0.3209 & 22.79 & 0.8540 & 0.1666 \\
          & LightenDiffusion$^\ast$ & 16.92   &0.5250  &  0.3824     & 18.27      & 0.7944      & 0.2457  \\
          \rowcolor{gray!30}
          & Ours  & 18.41 & 0.5341 & \textcolor{blue}{0.2974} & \textcolor{red}{21.32} & \textcolor{red}{0.8073} & \textcolor{blue}{0.1928} \\
          \rowcolor{gray!30}
          & Ours-LSRW & \textcolor{blue}{18.96} & 0.5438 & \textcolor{red}{0.2673} & 20.22 & 0.7649 & 0.2157 \\
    \midrule
    \end{tabular}%
    }
  \label{tab:ex_1}%
    \vspace{-2em}
\end{table}%

\subsection{Traditional Image Quality Comparison}
\textbf{Datasets} We use two low-light datasets LOLv2-real~\cite{yang2021sparse} and LSRW~\cite{hai2023r2rnet} captured in real environments for evaluation.
We report PSNR, SSIM, and LPIPS on both datasets.

\noindent \textbf{Results.} 
As shown in \cref{tab:ex_1}, LightenDiffusion~\cite{jiang2024lightendiffusion} performs best on LOL, but we note that it uses non-public data of 180k unpaired low/normal-light image pairs for training. For a fair comparison, we retrain it on the EnlightenGan~\cite{jiang2021enlightengan} dataset. More complete results, including the training sets used, are shown in the \textit{supplementary material}.

We can see that our method performs well on unseen data and the effects of traditional methods and supervised methods are not ideal.
Among them, supervised methods are trained on LOL, and they do not generalize well on LSRW. 
% RetinexNet\cite{wei2018deepretinexdecompositionlowlight}, Retinexformer\cite{cai2023retinexformer} and CIDNet\cite{yan2025hvi} a
Similarly, some unsupervised methods perform well on training-related datasets like NeRCo~\cite{yang2023implicit} and ZERO-IG~\cite{shi2024zero} but their performance drops significantly in unseen scenes. 
We also train our model on the LSRW, called Ours-LSRW, which also shows good generalization and significant performance improvements on LPIPS. 
% trained on LSRW and ZERO-IG\cite{shi2024zero} trained on LOL or LSRW,  On the contrary, our method performs well on unseen data, such as LSRW, which illustrates the superior generalization performance of our method.
% This indicates that our method effectively enhances the high-level semantic quality of generated images. 
Moreover, as shown in \cref{fig:ex_1}, our method obtains robust results under different lighting conditions, and some methods such as LightenDiffusion~\cite{jiang2024lightendiffusion}, Quadprior~\cite{wang2024zero} cannot handle lighting constraints and EnlightenGAN~\cite{jiang2021enlightengan}, NeRCo~\cite{yang2023implicit}, and Quadprior~\cite{wang2024zero} produce obvious artifacts.

We also note that although some methods like NeRCo~\cite{yang2023implicit} and LightenDiffusion~\cite{jiang2024lightendiffusion}, excel in image quality evaluation, their performance is significantly reduced for different high-level vision tasks. 
This indicates that visual quality is not strictly positively correlated with semantic quality, which is crucial for high-level vision.

% the luminous numbers in the first row, while some methods with lighting constraints such as EnlightenGAN\cite{jiang2021enlightengan}, Zero-DCE\cite{guo2020zero} cannot locally enhance the image brightness, resulting in image distortion in the second row. And 

\begin{table}[!t]
  \centering
  \caption{Ablation study on the proposed method.}
    \renewcommand{\arraystretch}{1.2}
    \resizebox{\linewidth}{!}{
    \begin{tabular}{c|c|c|cc|c|c|ccc|c|c}
    \toprule
    \multirow{2}[4]{*}{} & \multicolumn{1}{c|}{\multirow{2}[4]{*}{IA-IP}} & \multicolumn{1}{c|}{\multirow{2}[4]{*}{CC}}  & \multicolumn{2}{c|}{RC} & \multicolumn{1}{c|}{\multirow{2}[4]{*}{IPA}} & \multicolumn{1}{c|}{\multirow{2}[4]{*}{CAA}} & \multicolumn{3}{c|}{LSRW} & CODaN & FACE \\
   \cmidrule{4-5} \cmidrule{8-12}          & \multicolumn{1}{c|}{} & \multicolumn{1}{c|}{} & \multicolumn{1}{c|}{En} & \multicolumn{1}{c|}{De} & \multicolumn{1}{c|}{} & \multicolumn{1}{c|}{} & PSNR $\uparrow$ & SSIM $\uparrow$ & LPIPS $\downarrow$ & Top-1(\%) & mAP(\%) \\
    \midrule
    1     &  & & \multicolumn{1}{c|}{} &   & &   & 14.42 & 0.4683 & 0.3440 & 57.24 & 15.4 \\
    2     & \checkmark & & \multicolumn{1}{c|}{} &       & & & 17.99 & \textbf{0.5474} & 0.3223 & 59.40  & 15.5 \\
    3     & \checkmark & \checkmark & \multicolumn{1}{c|}{} &      & & & 16.98 & 0.5150 & 0.3279 & 59.92 & 16.0 \\
    4     & \checkmark &\checkmark & \multicolumn{1}{c|}{\checkmark} &      & &  & 18.48 & 0.5388 & 0.2989 & 60.04 & 15.4 \\ 
    5     & \checkmark & \checkmark & \multicolumn{1}{c|}{} & \checkmark    & &  & 18.48 & 0.5402 & 0.2916 & 60.52 & 16.2 \\
    6     & \checkmark & \checkmark & \multicolumn{1}{c|}{} & \checkmark    & \checkmark &  & 18.33 & 0.5412 & 0.2993 & 59.84 & 16.2 \\
    7     & \checkmark & \checkmark & \multicolumn{1}{c|}{} & \checkmark    &  & \checkmark & 18.41 & 0.5341 & 0.2974 & \textbf{60.92} & \textbf{16.9} \\
    8     & \checkmark & & \multicolumn{1}{c|}{} & \checkmark    &  & \checkmark  & \textbf{18.82} & 0.5410 & \textbf{0.2830} & 60.20  & 15.7 \\
    % \midrule
    % 9     & \multicolumn{4}{c|}{Baseline + Caption Prompt}  & 17.27 & 0.5278 & 0.3123 & 58.60  & 15.5 \\
    % 10     & \multicolumn{4}{c|}{Ours + Caption Prompt}  & 18.78 & 0.5460 & 0.2940 & 60.08 & 15.7\\
    \bottomrule
    \end{tabular}%
    }
  \label{tab:ab_1}%
  \vspace{-1.5em}
\end{table}%

\subsection{Model Analysis}
\begin{figure}[!t]
    \centering
      \includegraphics[width=1\linewidth]{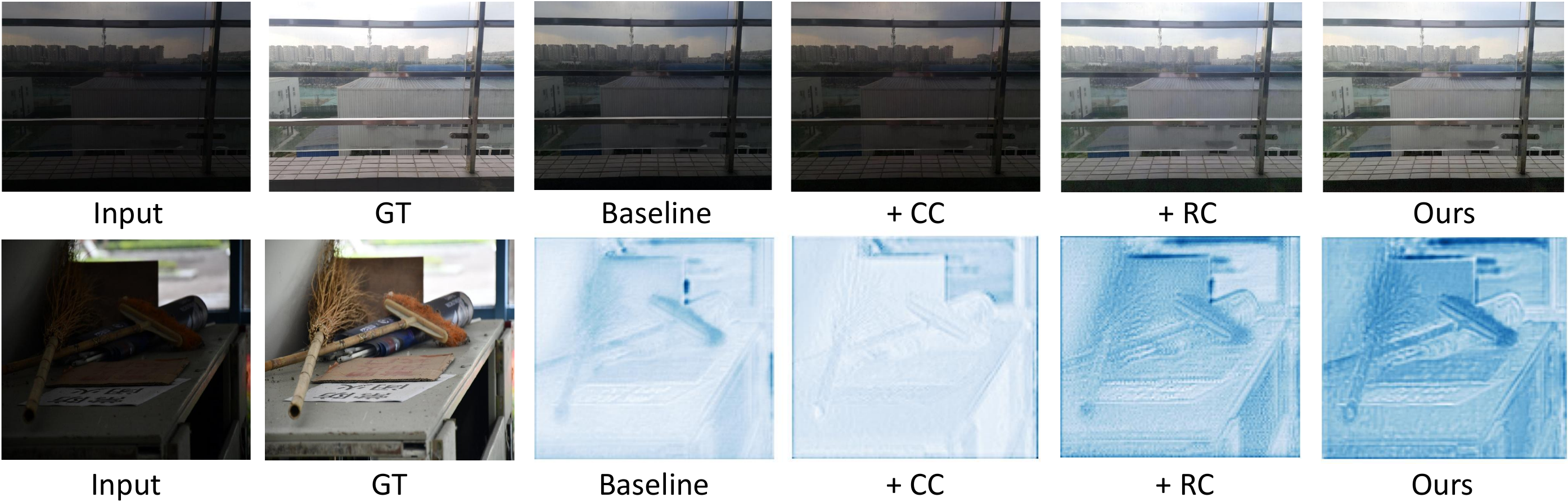}
    \caption{Visual analysis of the different components of the proposed method. CC denotes Caption Consistency, and RC denotes Reflectance Consistency.}
\label{fig:ab_1}
\vspace{-1.5em}
\end{figure}

\noindent \textbf{Ablation Study.} We comprehensively analyze the effectiveness of each component of the proposed framework in \cref{tab:ab_1}. 
Image translation model CycleGAN-Turbo~\cite{parmar2024one} is our baseline. 
To supplement the text prompt, we introduce the Illumination-Aware Image Prompt (IA-IP). 
% Unlike decoupled cross-attention\cite{ye2023ip}, 
We concatenate the text cross-attention and image cross-attention layers as a adapter. 
The results show significant improvements in both visual quality and downstream tasks ($2^{nd}$). 
The $3^{rd}$ row shows incorporating Caption Consistency (CC) enhances downstream performances. However, as shown in the first row of \cref{fig:ab_1}, without learning image-level information, the CC reduces the image visual quality.
% supporting our hypothesis that caption information boosts generalization but is less effective for pixel-level LIE tasks due to non-spatial features. 
To address this, we propose Reflectance Consistency (RC), which leverages an illumination-invariant reflectance map to learn robust spatial semantic features. 
The reflectance decoder can be connected after the lighten/darken encoder (En) or operate in parallel with the corresponding decoder (De). 
We explore decoder placement in $4^{th}$ and $5^{th}$ and the results show that De has better results, which allows the reflectance feature flow to pass through the encoder and the prompt adapter. 
In addition, we also compare the effects of different adapters including the original adapter ($5^{th}$), IP-Adapter (IPA, $6^{th}$), and the proposed CA-Adapter (CAA, $7^{th}$). The results show that the proposed CA-Adapter achieves the best performance. Furthermore, we observe that removing CC can yield better image visual quality but the generalization of the model in high-level vision tasks decreases as shown in the $8^{th}$ rows of \cref{tab:ab_1}.

% \noindent \textbf{Caption Prompt.} The caption prompt enriches the semantic description of the text prompt during the image enhancement process. A straightforward approach is to concatenate the caption with the text prompt and input them into the text encoder. As shown in \cref{tab:ab_2}, while this method improves both image visual quality and performance on high-level vision tasks, the cycle generation hinder the generalization of the caption and it is less efficient for inference with caption prompt. To address this, we propose Caption Consistency, a more efficient alternative that achieves superior generalization on downstream tasks. 

\begin{table}[!t]
  \centering
  \caption{Ablation study on caption prompt. For our method, we remove the caption consistency to evaluate directly using the caption prompt.}
    \renewcommand{\arraystretch}{1.2}
    \resizebox{\linewidth}{!}{
    \begin{tabular}{c|c|ccc|c|c}
    \toprule
    \multirow{2}[4]{*}{} & \multirow{2}[4]{*}{Method} & \multicolumn{3}{c|}{LSRW} & CODaN & Darkface \\
\cmidrule{3-7}       &   & PSNR $\uparrow$ & SSIM $\uparrow$ & LPIPS $\downarrow$ & Top-1(\%) & mAP(\%) \\
    \midrule
    1  & Baseline & 14.42 & 0.4683 & 0.3440 & 57.24 & 15.4 \\
    2 & Baseline + Caption Prompt & \textbf{17.27} & \textbf{0.5278} & \textbf{0.3123} & \textbf{58.60}  & \textbf{15.5} \\
    \midrule
    3 & Ours  & 18.41 & 0.5341 & 0.2974 & \textbf{60.92} & \textbf{16.9} \\
    4 & Ours + Caption Prompt & \textbf{18.78} & \textbf{0.5460} & \textbf{0.2940} & 60.08 & 15.7 \\
    \bottomrule
    \end{tabular}%
    }
  \label{tab:ab_2}%
    \vspace{-0.8em}
\end{table}%

\noindent \textbf{Caption Prompt.} The caption prompt has a more precise semantic description than the text prompt. A straightforward approach to enhancing the text prompt is to concatenate the caption with the text prompt and input them into the text encoder. As shown in \cref{tab:ab_2}, this method improves both image visual quality and performance on high-level vision tasks of baseline ($2^{nd}$), which shows the effectiveness of the caption prompt. However, the cycle generation process of our method hinders the generalization of the caption learning ($4^{th}$), achieving lower performances on downstream tasks. To address this, we propose Caption Consistency, a better alternative that achieves superior generalization on downstream tasks.

\noindent \textbf{Visual Analysis.} To demonstrate the effectiveness of our proposed method in improving the semantic quality, we visualize features of the last cross-attention layer of Unet in SD, as shown in the second row of \cref{fig:ab_1}. 
We present the results of the baseline and various components of the proposed method. Incorporating Caption Consistency enhances the distinction between different objects compared to the baseline, boosting performance on high-level vision tasks in the third row of \cref{tab:ab_1} but reducing attention to local details. Reflectance Consistency addresses this by raising attention to local details with a slight trade-off as shown in the fourth row of \cref{tab:ab_1}.
Additionally, our method employs a cycle-attention adapter to extract more expressive semantic features from illumination-aware image prompts, demonstrating a clear improvement in the semantic quality of the image compared to the baseline.

\noindent \textbf{Model complexity.} We utilize pretrained diffusion model for zero-shot capability. As shown in \cref{tab:ab_3}, 
we compare the input during inference, model size (with trainable parameters in parentheses), and the inference speed against other  methods based on pre-trained diffusion models.  Our fine-tuning method achieves the best efficiency. Additionally, due to the complex input and model design of LightDiff~\cite{li2024light}, it does not exhibit zero-shot performance, so we exclude its results in the quantitative comparison.

\begin{table}[!t]
  \centering
  \caption{Model complexity of other low-light enhancement methods based on pre-trained diffusion models, including the input during inference, model size, and inference time.}
    \renewcommand{\arraystretch}{1.2}
    \scalebox{0.8}{
    \begin{tabular}{c|ccccc}
    \hline
    Method & Text  & Caption & Depth  & SIZE (M) & TIME (S) \\
    \hline
    QuadPrior\cite{wang2024zero} &  &  &  &1314(1232.5) & 2.443 \\
    LightDiff\cite{li2024light} & \checkmark & \checkmark & \checkmark &1670(1220.1) & 12.138\\
    Ours   &\checkmark  &  &  & \textbf{598}(\textbf{1.7})  & \textbf{0.348} \\
    \hline
    \end{tabular}%
    }
  \label{tab:ab_3}%
      \vspace{-1.5em}
\end{table}%
\section{Conclusion}
In this paper, we design a benchmark GEFU for low-light vision and a semantically consistent unsupervised fine-tuning framework.
Our approach includes an illumination-aware image prompt that incorporates a cycle-attention adapter, caption consistency, and reflectance consistency and demonstrates superior performance.

\noindent  \textbf{Future Work.} Our experiments reveal that visual quality and performance in high-level vision tasks are not strictly positively correlated, highlighting the need for further exploration of their relationship. Additionally, designing lightweight networks using techniques such as knowledge distillation can enhance real-time performance and facilitate their application to downstream tasks.

\noindent  \textbf{Acknowledgment.} This work is supported by the National Natural Science Foundation of China No.62302167, U23A20343, 62222602, 62176092, 62476090, 72192821; Shanghai Sailing Program 23YF1410500; Young Elite Scientists Sponsorship Program by CAST YESS20240780; the Chenguang Program of Shanghai Education Development Foundation and Shanghai Municipal Education Commission 23CGA34; Natural Science Foundation of Chongqing CSTB2023NSCQJQX0007, CSTB2023NSCQ-MSX0137; CCF-Tencent RAGR20240122, RAGR20230121; the Open Research Fund of Key Laboratory of Advanced Theory and Application in Statistics and Data Science-MOE, ECNU.
{
    \small
    \bibliographystyle{ieeenat_fullname}
    \bibliography{main}

\begin{thebibliography}{70}
\providecommand{\natexlab}[1]{#1}
\providecommand{\url}[1]{\texttt{#1}}
\expandafter\ifx\csname urlstyle\endcsname\relax
  \providecommand{\doi}[1]{doi: #1}\else
  \providecommand{\doi}{doi: \begingroup \urlstyle{rm}\Url}\fi

\bibitem[Bovik(2010)]{bovik2010handbook}
Alan~C Bovik.
\newblock \emph{Handbook of image and video processing}.
\newblock Academic press, 2010.

\bibitem[Cai et~al.(2023)Cai, Bian, Lin, Wang, Timofte, and Zhang]{cai2023retinexformer}
Yuanhao Cai, Hao Bian, Jing Lin, Haoqian Wang, Radu Timofte, and Yulun Zhang.
\newblock Retinexformer: One-stage retinex-based transformer for low-light image enhancement.
\newblock In \emph{Proceedings of the IEEE/CVF International Conference on Computer Vision}, pages 12504--12513, 2023.

\bibitem[Cordts et~al.(2016)Cordts, Omran, Ramos, Rehfeld, Enzweiler, Benenson, Franke, Roth, and Schiele]{cordts2016cityscapes}
Marius Cordts, Mohamed Omran, Sebastian Ramos, Timo Rehfeld, Markus Enzweiler, Rodrigo Benenson, Uwe Franke, Stefan Roth, and Bernt Schiele.
\newblock The cityscapes dataset for semantic urban scene understanding.
\newblock In \emph{Proceedings of the IEEE conference on computer vision and pattern recognition}, pages 3213--3223, 2016.

\bibitem[Cui et~al.(2021)Cui, Qi, Gu, You, Zhang, and Harada]{cui2021multitask}
Ziteng Cui, Guo-Jun Qi, Lin Gu, Shaodi You, Zenghui Zhang, and Tatsuya Harada.
\newblock Multitask aet with orthogonal tangent regularity for dark object detection.
\newblock In \emph{Proceedings of the IEEE/CVF international conference on computer vision}, pages 2553--2562, 2021.

\bibitem[Cui et~al.(2022)Cui, Li, Gu, Su, Gao, Jiang, Qiao, and Harada]{Cui_2022_BMVC}
Ziteng Cui, Kunchang Li, Lin Gu, Shenghan Su, Peng Gao, ZhengKai Jiang, Yu Qiao, and Tatsuya Harada.
\newblock You only need 90k parameters to adapt light: a light weight transformer for image enhancement and exposure correction.
\newblock In \emph{33rd British Machine Vision Conference 2022, {BMVC} 2022, London, UK, November 21-24, 2022}. {BMVA} Press, 2022.

\bibitem[Du et~al.(2024)Du, Shi, and Deng]{du2024boosting}
Zhipeng Du, Miaojing Shi, and Jiankang Deng.
\newblock Boosting object detection with zero-shot day-night domain adaptation.
\newblock In \emph{Proceedings of the IEEE/CVF Conference on Computer Vision and Pattern Recognition}, pages 12666--12676, 2024.

\bibitem[Farhadi and Redmon(2018)]{farhadi2018yolov3}
Ali Farhadi and Joseph Redmon.
\newblock Yolov3: An incremental improvement.
\newblock In \emph{Computer vision and pattern recognition}, pages 1--6. Springer Berlin/Heidelberg, Germany, 2018.

\bibitem[Fu et~al.(2023)Fu, Yang, Tu, Huang, Ding, and Ma]{fu2023learning}
Zhenqi Fu, Yan Yang, Xiaotong Tu, Yue Huang, Xinghao Ding, and Kai-Kuang Ma.
\newblock Learning a simple low-light image enhancer from paired low-light instances.
\newblock In \emph{Proceedings of the IEEE/CVF conference on computer vision and pattern recognition}, pages 22252--22261, 2023.

\bibitem[Gong et~al.(2023)Gong, Lou, Liu, Zhang, Chen, Zhang, Tan, Xie, and Ma]{Gong2023}
Jingyu Gong, Yujing Lou, Fengqi Liu, Zhiwei Zhang, Haoming Chen, Zhizhong Zhang, Xin Tan, Yuan Xie, and Lizhuang Ma.
\newblock Scene point cloud understanding and reconstruction technologies in 3d space.
\newblock \emph{Journal of Image and Graphics}, 28\penalty0 (06):\penalty0 1741--1766, 2023.

\bibitem[Gonzales and Wintz(1987)]{gonzales1987digital}
Rafael~C Gonzales and Paul Wintz.
\newblock \emph{Digital image processing}.
\newblock Addison-Wesley Longman Publishing Co., Inc., 1987.

\bibitem[Guo et~al.(2020)Guo, Li, Guo, Loy, Hou, Kwong, and Cong]{guo2020zero}
Chunle Guo, Chongyi Li, Jichang Guo, Chen~Change Loy, Junhui Hou, Sam Kwong, and Runmin Cong.
\newblock Zero-reference deep curve estimation for low-light image enhancement.
\newblock In \emph{Proceedings of the IEEE/CVF conference on computer vision and pattern recognition}, pages 1780--1789, 2020.

\bibitem[Guo et~al.(2016)Guo, Li, and Ling]{guo2016lime}
Xiaojie Guo, Yu Li, and Haibin Ling.
\newblock Lime: Low-light image enhancement via illumination map estimation.
\newblock \emph{IEEE Transactions on image processing}, 26\penalty0 (2):\penalty0 982--993, 2016.

\bibitem[Hai et~al.(2023)Hai, Xuan, Yang, Hao, Zou, Lin, and Han]{hai2023r2rnet}
Jiang Hai, Zhu Xuan, Ren Yang, Yutong Hao, Fengzhu Zou, Fang Lin, and Songchen Han.
\newblock R2rnet: Low-light image enhancement via real-low to real-normal network.
\newblock \emph{Journal of Visual Communication and Image Representation}, 90:\penalty0 103712, 2023.

\bibitem[He et~al.(2016)He, Zhang, Ren, and Sun]{he2016deep}
Kaiming He, Xiangyu Zhang, Shaoqing Ren, and Jian Sun.
\newblock Deep residual learning for image recognition.
\newblock In \emph{Proceedings of the IEEE conference on computer vision and pattern recognition}, pages 770--778, 2016.

\bibitem[Hoyer et~al.(2022{\natexlab{a}})Hoyer, Dai, and Van~Gool]{hoyer2022daformer}
Lukas Hoyer, Dengxin Dai, and Luc Van~Gool.
\newblock Daformer: Improving network architectures and training strategies for domain-adaptive semantic segmentation.
\newblock In \emph{Proceedings of the IEEE/CVF conference on computer vision and pattern recognition}, pages 9924--9935, 2022{\natexlab{a}}.

\bibitem[Hoyer et~al.(2022{\natexlab{b}})Hoyer, Dai, and Van~Gool]{hoyer2022hrda}
Lukas Hoyer, Dengxin Dai, and Luc Van~Gool.
\newblock Hrda: Context-aware high-resolution domain-adaptive semantic segmentation.
\newblock In \emph{European conference on computer vision}, pages 372--391. Springer, 2022{\natexlab{b}}.

\bibitem[Hu et~al.(2021)Hu, Shen, Wallis, Allen-Zhu, Li, Wang, Wang, and Chen]{hu2021lora}
Edward~J Hu, Yelong Shen, Phillip Wallis, Zeyuan Allen-Zhu, Yuanzhi Li, Shean Wang, Lu Wang, and Weizhu Chen.
\newblock Lora: Low-rank adaptation of large language models.
\newblock \emph{arXiv preprint arXiv:2106.09685}, 2021.

\bibitem[Hu et~al.(2022)Hu, Shen, Wallis, Allen-Zhu, Li, Wang, Wang, Chen, et~al.]{hu2022lora}
Edward~J Hu, Yelong Shen, Phillip Wallis, Zeyuan Allen-Zhu, Yuanzhi Li, Shean Wang, Lu Wang, Weizhu Chen, et~al.
\newblock Lora: Low-rank adaptation of large language models.
\newblock \emph{ICLR}, 1\penalty0 (2):\penalty0 3, 2022.

\bibitem[Jiang et~al.(2024)Jiang, Luo, Liu, Han, and Liu]{jiang2024lightendiffusion}
Hai Jiang, Ao Luo, Xiaohong Liu, Songchen Han, and Shuaicheng Liu.
\newblock Lightendiffusion: Unsupervised low-light image enhancement with latent-retinex diffusion models.
\newblock \emph{arXiv preprint arXiv:2407.08939}, 2024.

\bibitem[Jiang et~al.(2021)Jiang, Gong, Liu, Cheng, Fang, Shen, Yang, Zhou, and Wang]{jiang2021enlightengan}
Yifan Jiang, Xinyu Gong, Ding Liu, Yu Cheng, Chen Fang, Xiaohui Shen, Jianchao Yang, Pan Zhou, and Zhangyang Wang.
\newblock Enlightengan: Deep light enhancement without paired supervision.
\newblock \emph{IEEE transactions on image processing}, 30:\penalty0 2340--2349, 2021.

\bibitem[Kumari et~al.(2022)Kumari, Zhang, Shechtman, and Zhu]{kumari2022ensembling}
Nupur Kumari, Richard Zhang, Eli Shechtman, and Jun-Yan Zhu.
\newblock Ensembling off-the-shelf models for gan training.
\newblock In \emph{Proceedings of the IEEE/CVF conference on computer vision and pattern recognition}, pages 10651--10662, 2022.

\bibitem[Land(1977)]{land1977retinex}
Edwin~H Land.
\newblock The retinex theory of color vision.
\newblock \emph{Scientific american}, 237\penalty0 (6):\penalty0 108--129, 1977.

\bibitem[Lengyel et~al.(2021)Lengyel, Garg, Milford, and van Gemert]{lengyel2021zero}
Attila Lengyel, Sourav Garg, Michael Milford, and Jan~C van Gemert.
\newblock Zero-shot day-night domain adaptation with a physics prior.
\newblock In \emph{Proceedings of the IEEE/CVF International Conference on Computer Vision}, pages 4399--4409, 2021.

\bibitem[Li et~al.(2021)Li, Guo, and Loy]{li2021learning}
Chongyi Li, Chunle Guo, and Chen~Change Loy.
\newblock Learning to enhance low-light image via zero-reference deep curve estimation.
\newblock \emph{IEEE transactions on pattern analysis and machine intelligence}, 44\penalty0 (8):\penalty0 4225--4238, 2021.

\bibitem[Li et~al.(2019)Li, Wang, Wang, Tai, Qian, Yang, Wang, Li, and Huang]{li2019dsfd}
Jian Li, Yabiao Wang, Changan Wang, Ying Tai, Jianjun Qian, Jian Yang, Chengjie Wang, Jilin Li, and Feiyue Huang.
\newblock Dsfd: dual shot face detector.
\newblock In \emph{Proceedings of the IEEE/CVF conference on computer vision and pattern recognition}, pages 5060--5069, 2019.

\bibitem[Li et~al.(2022)Li, Li, Xiong, and Hoi]{li2022blip}
Junnan Li, Dongxu Li, Caiming Xiong, and Steven Hoi.
\newblock Blip: Bootstrapping language-image pre-training for unified vision-language understanding and generation.
\newblock In \emph{International conference on machine learning}, pages 12888--12900. PMLR, 2022.

\bibitem[Li et~al.(2024)Li, Li, Tu, Liu, Guo, Juefei-Xu, Xu, and Yu]{li2024light}
Jinlong Li, Baolu Li, Zhengzhong Tu, Xinyu Liu, Qing Guo, Felix Juefei-Xu, Runsheng Xu, and Hongkai Yu.
\newblock Light the night: A multi-condition diffusion framework for unpaired low-light enhancement in autonomous driving.
\newblock In \emph{Proceedings of the IEEE/CVF Conference on Computer Vision and Pattern Recognition}, pages 15205--15215, 2024.

\bibitem[Liang et~al.(2023)Liang, Li, Zhou, Feng, and Loy]{liang2023iterative}
Zhexin Liang, Chongyi Li, Shangchen Zhou, Ruicheng Feng, and Chen~Change Loy.
\newblock Iterative prompt learning for unsupervised backlit image enhancement.
\newblock In \emph{Proceedings of the IEEE/CVF International Conference on Computer Vision}, pages 8094--8103, 2023.

\bibitem[Lin et~al.(2017)Lin, Milan, Shen, and Reid]{lin2017refinenet}
Guosheng Lin, Anton Milan, Chunhua Shen, and Ian Reid.
\newblock Refinenet: Multi-path refinement networks for high-resolution semantic segmentation.
\newblock In \emph{Proceedings of the IEEE conference on computer vision and pattern recognition}, pages 1925--1934, 2017.

\bibitem[Liu et~al.(2021)Liu, Ma, Zhang, Fan, and Luo]{liu2021retinex}
Risheng Liu, Long Ma, Jiaao Zhang, Xin Fan, and Zhongxuan Luo.
\newblock Retinex-inspired unrolling with cooperative prior architecture search for low-light image enhancement.
\newblock In \emph{Proceedings of the IEEE/CVF conference on computer vision and pattern recognition}, pages 10561--10570, 2021.

\bibitem[Loh and Chan(2019)]{loh2019getting}
Yuen~Peng Loh and Chee~Seng Chan.
\newblock Getting to know low-light images with the exclusively dark dataset.
\newblock \emph{Computer Vision and Image Understanding}, 178:\penalty0 30--42, 2019.

\bibitem[Loshchilov and Hutter(2019)]{loshchilov2018decoupled}
Ilya Loshchilov and Frank Hutter.
\newblock Decoupled weight decay regularization.
\newblock In \emph{International Conference on Learning Representations}, 2019.

\bibitem[Luo et~al.(2023)Luo, Wang, Yang, and Liu]{luo2023similarity}
Rundong Luo, Wenjing Wang, Wenhan Yang, and Jiaying Liu.
\newblock Similarity min-max: Zero-shot day-night domain adaptation.
\newblock In \emph{Proceedings of the IEEE/CVF International Conference on Computer Vision}, pages 8104--8114, 2023.

\bibitem[Ma et~al.(2024)Ma, Yang, and Huang]{ma2024taming}
Hao Ma, Jingyuan Yang, and Hui Huang.
\newblock Taming diffusion model for exemplar-based image translation.
\newblock \emph{Computational Visual Media}, 10\penalty0 (6):\penalty0 1031--1043, 2024.

\bibitem[Ma et~al.(2022)Ma, Ma, Liu, Fan, and Luo]{ma2022toward}
Long Ma, Tengyu Ma, Risheng Liu, Xin Fan, and Zhongxuan Luo.
\newblock Toward fast, flexible, and robust low-light image enhancement.
\newblock In \emph{Proceedings of the IEEE/CVF conference on computer vision and pattern recognition}, pages 5637--5646, 2022.

\bibitem[Park et~al.(2022)Park, Vien, Kim, and Lee]{park2022histogram}
Jaemin Park, An~Gia Vien, Jin-Hwan Kim, and Chul Lee.
\newblock Histogram-based transformation function estimation for low-light image enhancement.
\newblock In \emph{2022 IEEE International Conference on Image Processing (ICIP)}, pages 1--5. IEEE, 2022.

\bibitem[Parmar et~al.(2024)Parmar, Park, Narasimhan, and Zhu]{parmar2024one}
Gaurav Parmar, Taesung Park, Srinivasa Narasimhan, and Jun-Yan Zhu.
\newblock One-step image translation with text-to-image models.
\newblock \emph{arXiv preprint arXiv:2403.12036}, 2024.

\bibitem[Peng et~al.(2024)Peng, Zhu, Jiang, Tai, Luo, Zhang, Lin, Jin, Wang, and Ji]{peng2024portraitbooth}
Xu Peng, Junwei Zhu, Boyuan Jiang, Ying Tai, Donghao Luo, Jiangning Zhang, Wei Lin, Taisong Jin, Chengjie Wang, and Rongrong Ji.
\newblock Portraitbooth: A versatile portrait model for fast identity-preserved personalization.
\newblock In \emph{Proceedings of the IEEE/CVF Conference on Computer Vision and Pattern Recognition}, pages 27080--27090, 2024.

\bibitem[Radford et~al.(2021)Radford, Kim, Hallacy, Ramesh, Goh, Agarwal, Sastry, Askell, Mishkin, Clark, Krueger, and Sutskever]{pmlr-clip}
Alec Radford, Jong~Wook Kim, Chris Hallacy, Aditya Ramesh, Gabriel Goh, Sandhini Agarwal, Girish Sastry, Amanda Askell, Pamela Mishkin, Jack Clark, Gretchen Krueger, and Ilya Sutskever.
\newblock Learning transferable visual models from natural language supervision.
\newblock In \emph{Proceedings of the 38th International Conference on Machine Learning}, pages 8748--8763. PMLR, 2021.

\bibitem[Ramesh et~al.(2022)Ramesh, Dhariwal, Nichol, Chu, and Chen]{ramesh2022hierarchical}
Aditya Ramesh, Prafulla Dhariwal, Alex Nichol, Casey Chu, and Mark Chen.
\newblock Hierarchical text-conditional image generation with clip latents.
\newblock \emph{arXiv preprint arXiv:2204.06125}, 1\penalty0 (2):\penalty0 3, 2022.

\bibitem[Rombach et~al.(2022{\natexlab{a}})Rombach, Blattmann, Lorenz, Esser, and Ommer]{Rombach_2022_CVPR}
Robin Rombach, Andreas Blattmann, Dominik Lorenz, Patrick Esser, and Bj\"orn Ommer.
\newblock High-resolution image synthesis with latent diffusion models.
\newblock In \emph{Proceedings of the IEEE/CVF Conference on Computer Vision and Pattern Recognition (CVPR)}, pages 10684--10695, 2022{\natexlab{a}}.

\bibitem[Rombach et~al.(2022{\natexlab{b}})Rombach, Blattmann, Lorenz, Esser, and Ommer]{rombach2022high}
Robin Rombach, Andreas Blattmann, Dominik Lorenz, Patrick Esser, and Bj{\"o}rn Ommer.
\newblock High-resolution image synthesis with latent diffusion models.
\newblock In \emph{Proceedings of the IEEE/CVF conference on computer vision and pattern recognition}, pages 10684--10695, 2022{\natexlab{b}}.

\bibitem[Ruiz et~al.(2023)Ruiz, Li, Jampani, Pritch, Rubinstein, and Aberman]{ruiz2023dreambooth}
Nataniel Ruiz, Yuanzhen Li, Varun Jampani, Yael Pritch, Michael Rubinstein, and Kfir Aberman.
\newblock Dreambooth: Fine tuning text-to-image diffusion models for subject-driven generation.
\newblock In \emph{Proceedings of the IEEE/CVF conference on computer vision and pattern recognition}, pages 22500--22510, 2023.

\bibitem[Saharia et~al.(2022)Saharia, Chan, Saxena, Li, Whang, Denton, Ghasemipour, Gontijo~Lopes, Karagol~Ayan, Salimans, et~al.]{saharia2022photorealistic}
Chitwan Saharia, William Chan, Saurabh Saxena, Lala Li, Jay Whang, Emily~L Denton, Kamyar Ghasemipour, Raphael Gontijo~Lopes, Burcu Karagol~Ayan, Tim Salimans, et~al.
\newblock Photorealistic text-to-image diffusion models with deep language understanding.
\newblock \emph{Advances in neural information processing systems}, 35:\penalty0 36479--36494, 2022.

\bibitem[Sakaridis et~al.(2019)Sakaridis, Dai, and Gool]{sakaridis2019guided}
Christos Sakaridis, Dengxin Dai, and Luc~Van Gool.
\newblock Guided curriculum model adaptation and uncertainty-aware evaluation for semantic nighttime image segmentation.
\newblock In \emph{Proceedings of the IEEE/CVF international conference on computer vision}, pages 7374--7383, 2019.

\bibitem[Sauer et~al.(2023)Sauer, Lorenz, Blattmann, and Rombach]{sauer2023adversarialdiffusiondistillation}
Axel Sauer, Dominik Lorenz, Andreas Blattmann, and Robin Rombach.
\newblock Adversarial diffusion distillation, 2023.

\bibitem[Sauer et~al.(2024)Sauer, Lorenz, Blattmann, and Rombach]{sauer2024adversarial}
Axel Sauer, Dominik Lorenz, Andreas Blattmann, and Robin Rombach.
\newblock Adversarial diffusion distillation.
\newblock In \emph{European Conference on Computer Vision}, pages 87--103. Springer, 2024.

\bibitem[Shi et~al.(2024)Shi, Liu, Zhang, Tian, Xia, and Fu]{shi2024zero}
Yiqi Shi, Duo Liu, Liguo Zhang, Ye Tian, Xuezhi Xia, and Xiaojing Fu.
\newblock Zero-ig: Zero-shot illumination-guided joint denoising and adaptive enhancement for low-light images.
\newblock In \emph{Proceedings of the IEEE/CVF Conference on Computer Vision and Pattern Recognition}, pages 3015--3024, 2024.

\bibitem[Tan et~al.(2021)Tan, Xu, Cao, Zhang, Ma, and Lau]{Tan_2021_TIP_NightCity}
Xin Tan, Ke Xu, Ying Cao, Yiheng Zhang, Lizhuang Ma, and Rynson W.~H. Lau.
\newblock Night-time scene parsing with a large real dataset.
\newblock \emph{IEEE Transactions on Image Processing}, 30:\penalty0 9085--9098, 2021.

\bibitem[Wang et~al.(2021)Wang, Yang, and Liu]{wang2021hla}
Wenjing Wang, Wenhan Yang, and Jiaying Liu.
\newblock Hla-face: Joint high-low adaptation for low light face detection.
\newblock In \emph{Proceedings of the IEEE/CVF Conference on Computer Vision and Pattern Recognition}, pages 16195--16204, 2021.

\bibitem[Wang et~al.(2024{\natexlab{a}})Wang, Liu, Zhou, Wei, Deng, Murshed, and Lu]{wang2024noise4denoise}
Weijia Wang, Xiao Liu, Hailing Zhou, Lei Wei, Zhigang Deng, Manzur Murshed, and Xuequan Lu.
\newblock Noise4denoise: Leveraging noise for unsupervised point cloud denoising.
\newblock \emph{Computational Visual Media}, 10\penalty0 (4):\penalty0 659--669, 2024{\natexlab{a}}.

\bibitem[Wang et~al.(2024{\natexlab{b}})Wang, Luo, Yang, and Liu]{wang2024unsupervised}
Wenjing Wang, Rundong Luo, Wenhan Yang, and Jiaying Liu.
\newblock Unsupervised illumination adaptation for low-light vision.
\newblock \emph{IEEE Transactions on Pattern Analysis \& Machine Intelligence}, \penalty0 (01):\penalty0 1--15, 2024{\natexlab{b}}.

\bibitem[Wang et~al.(2024{\natexlab{c}})Wang, Yang, Fu, and Liu]{wang2024zero}
Wenjing Wang, Huan Yang, Jianlong Fu, and Jiaying Liu.
\newblock Zero-reference low-light enhancement via physical quadruple priors.
\newblock In \emph{Proceedings of the IEEE/CVF Conference on Computer Vision and Pattern Recognition}, pages 26057--26066, 2024{\natexlab{c}}.

\bibitem[Wei et~al.(2018)Wei, Wang, Yang, and Liu]{wei2018deepretinexdecompositionlowlight}
Chen Wei, Wenjing Wang, Wenhan Yang, and Jiaying Liu.
\newblock Deep retinex decomposition for low-light enhancement, 2018.

\bibitem[Wu et~al.(2023)Wu, Pan, Wang, Yang, Wei, Li, and Shen]{wu2023learning}
Yuhui Wu, Chen Pan, Guoqing Wang, Yang Yang, Jiwei Wei, Chongyi Li, and Heng~Tao Shen.
\newblock Learning semantic-aware knowledge guidance for low-light image enhancement.
\newblock In \emph{Proceedings of the IEEE/CVF Conference on Computer Vision and Pattern Recognition}, pages 1662--1671, 2023.

\bibitem[Xie et~al.(2023)Xie, Wang, Xu, Zhang, Tan, Xie, and Ma]{wang_fdlnet}
Zhifeng Xie, Sen Wang, Ke Xu, Zhizhong Zhang, Xin Tan, Yuan Xie, and Lizhuang Ma.
\newblock Boosting night-time scene parsing with learnable frequency.
\newblock \emph{IEEE Transactions on Image Processing}, 32:\penalty0 2386--2398, 2023.

\bibitem[Xie et~al.(2024)Xie, Qiu, Wang, Tan, Xie, and Ma]{rui_pig}
Zhifeng Xie, Rui Qiu, Sen Wang, Xin Tan, Yuan Xie, and Lizhuang Ma.
\newblock Pig: Prompt images guidance for night-time scene parsing.
\newblock \emph{IEEE Transactions on Image Processing}, 33:\penalty0 3921--3934, 2024.

\bibitem[Yan et~al.(2025)Yan, Feng, Zhang, Pang, Shi, Wu, Dong, Sun, and Zhang]{yan2025hvi}
Qingsen Yan, Yixu Feng, Cheng Zhang, Guansong Pang, Kangbiao Shi, Peng Wu, Wei Dong, Jinqiu Sun, and Yanning Zhang.
\newblock Hvi: A new color space for low-light image enhancement.
\newblock \emph{arXiv preprint arXiv:2502.20272}, 2025.

\bibitem[Yan et~al.(2024)Yan, Zheng, Fan, Li, Li, and Yang]{yan2024learnable}
Zhiqiang Yan, Yupeng Zheng, Deng-Ping Fan, Xiang Li, Jun Li, and Jian Yang.
\newblock Learnable differencing center for nighttime depth perception.
\newblock \emph{Visual Intelligence}, 2\penalty0 (1):\penalty0 15, 2024.

\bibitem[Yang et~al.(2016)Yang, Luo, Loy, and Tang]{yang2016wider}
Shuo Yang, Ping Luo, Chen-Change Loy, and Xiaoou Tang.
\newblock Wider face: A face detection benchmark.
\newblock In \emph{Proceedings of the IEEE conference on computer vision and pattern recognition}, pages 5525--5533, 2016.

\bibitem[Yang et~al.(2023)Yang, Ding, Wu, Li, and Zhang]{yang2023implicit}
Shuzhou Yang, Moxuan Ding, Yanmin Wu, Zihan Li, and Jian Zhang.
\newblock Implicit neural representation for cooperative low-light image enhancement.
\newblock In \emph{Proceedings of the IEEE/CVF international conference on computer vision}, pages 12918--12927, 2023.

\bibitem[Yang et~al.(2020)Yang, Yuan, Ren, Liu, Scheirer, Wang, , and et~al.]{poor_visibility_benchmark}
Wenhan Yang, Ye Yuan, Wenqi Ren, Jiaying Liu, Walter~J. Scheirer, Zhangyang Wang, , and et al.
\newblock Advancing image understanding in poor visibility environments: A collective benchmark study.
\newblock \emph{IEEE Transactions on Image Processing}, 29:\penalty0 5737--5752, 2020.

\bibitem[Yang et~al.(2021)Yang, Wang, Huang, Wang, and Liu]{yang2021sparse}
Wenhan Yang, Wenjing Wang, Haofeng Huang, Shiqi Wang, and Jiaying Liu.
\newblock Sparse gradient regularized deep retinex network for robust low-light image enhancement.
\newblock \emph{IEEE Transactions on Image Processing}, 30:\penalty0 2072--2086, 2021.

\bibitem[Ye et~al.(2023)Ye, Zhang, Liu, Han, and Yang]{ye2023ip}
Hu Ye, Jun Zhang, Sibo Liu, Xiao Han, and Wei Yang.
\newblock Ip-adapter: Text compatible image prompt adapter for text-to-image diffusion models.
\newblock \emph{arXiv preprint arXiv:2308.06721}, 2023.

\bibitem[Yu et~al.(2020)Yu, Chen, Wang, Xian, Chen, Liu, Madhavan, and Darrell]{yu2020bdd100k}
Fisher Yu, Haofeng Chen, Xin Wang, Wenqi Xian, Yingying Chen, Fangchen Liu, Vashisht Madhavan, and Trevor Darrell.
\newblock Bdd100k: A diverse driving dataset for heterogeneous multitask learning.
\newblock In \emph{Proceedings of the IEEE/CVF conference on computer vision and pattern recognition}, pages 2636--2645, 2020.

\bibitem[Zhang et~al.(2019{\natexlab{a}})Zhang, Nie, and Zheng]{zhang2019dual}
Qing Zhang, Yongwei Nie, and Wei-Shi Zheng.
\newblock Dual illumination estimation for robust exposure correction.
\newblock In \emph{Computer graphics forum}, pages 243--252. Wiley Online Library, 2019{\natexlab{a}}.

\bibitem[Zhang et~al.(2019{\natexlab{b}})Zhang, Zhang, and Guo]{zhang2019kindling}
Yonghua Zhang, Jiawan Zhang, and Xiaojie Guo.
\newblock Kindling the darkness: A practical low-light image enhancer.
\newblock In \emph{Proceedings of the 27th ACM international conference on multimedia}, pages 1632--1640, 2019{\natexlab{b}}.

\bibitem[Zhao et~al.(2022)Zhao, Zhang, Lu, Li, Wu, and Sheng]{zhao2022dsd}
Yicheng Zhao, Han Zhang, Ping Lu, Ping Li, Enhua Wu, and Bin Sheng.
\newblock Dsd-matchingnet: Deformable sparse-to-dense feature matching for learning accurate correspondences.
\newblock \emph{Virtual Reality \& Intelligent Hardware}, 4\penalty0 (5):\penalty0 432--443, 2022.

\bibitem[Zheng et~al.(2023)Zheng, Huang, Zhou, Yang, Zhu, and Zhao]{zheng2023learning}
Naishan Zheng, Jie Huang, Man Zhou, Zizheng Yang, Qi Zhu, and Feng Zhao.
\newblock Learning semantic degradation-aware guidance for recognition-driven unsupervised low-light image enhancement.
\newblock In \emph{Proceedings of the AAAI Conference on Artificial Intelligence}, pages 3678--3686, 2023.

\bibitem[Zhu et~al.(2017)Zhu, Park, Isola, and Efros]{zhu2017unpaired}
Jun-Yan Zhu, Taesung Park, Phillip Isola, and Alexei~A Efros.
\newblock Unpaired image-to-image translation using cycle-consistent adversarial networks.
\newblock In \emph{Proceedings of the IEEE international conference on computer vision}, pages 2223--2232, 2017.

\end{thebibliography}
}
% \clearpage
\setcounter{page}{1}
\renewcommand\thesection{\Alph{section}}
\setcounter{section}{0}

\onecolumn % 切换到单栏模式
% \maketitle
% \tableofcontents

\clearpage
\section{Implementation details}
\label{sec:ex_d}

\subsection{Hyper Parameters}
We use the AdamW optimizer to train our model, the weight decay and epsilon are set to 1e-2 and 1e-8, respectively. During training, the weights of $\lambda_{idt}$ and $\lambda_{GAN}$ are 0.5 and 1, respectively, we use gradient clipping with a max norm of 10, and low light and normal light images are randomly paired to ensure the generalization of the model. 

\subsection{Algorithm Flow}
We use pseudo code as shown in \cref{alg:al_1} to illustrate the process of our method more fully, and \cref{alg:al_2} describes in more detail the specific process of the proposed caption consistency and reflectance consistency.

\begin{algorithm}[ht]
\caption{Pipeline of the Proposed Method SCUF}
\label{alg:al_1}
\begin{algorithmic}[1]
\STATE \textbf{Input:} 

(1) the low-light image $I_l$ and normal-light image $I_n$ 

(2) the V channel image $I_{l,v}$, $I_{n,v}$ and reverse image $I_{l,v}^{r}$, $I_{n,v}^{r}$ from $I_l$ and $I_n$ in HSV color space, respectively.

(3) the text prompt $T_l$ and $T_d$ for lightening and darkening, respectively.

\STATE \textbf{Networks:} 
The lighten encoder $E_{l}$ and decoder $D_{l}$, the draken encoder $E_{d}$ and decoder $D_{d}$, and initial fixed Unet $U$, lightening and darkening discriminators ${Dis}_l$ and ${Dis}_n$

\FOR{$i$ in $1:iterations$}
\FOR {\textbf{the cycle generation}}

\STATE Input $I_l$ and obtain the generated normal-light image $\hat{I}_n$ and low-light image $I_l^{'}$ by:

$\hat{I}_n = D_{l}(U(E_l(I_l),(T_l,I_{l,v}^{r})))$

$I_l^{'} = D_{d}(U(E_d(\hat{I}_n),(T_d,I_{l,v})))$

\STATE \textbf{Do caption and reflectance consistency}

\STATE Input $I_n$ and obtain the generated low-light image $\hat{I}_l$ and normal-light image $I_n^{'}$ by:

$\hat{I}_l = D_{d}(U(E_d(I_n),(T_d,I_{n,v}^{r})))$

$I_n^{'} = D_{l}(U(E_l(\hat{I}_l),(T_l,I_{n,v})))$

\STATE \textbf{Do caption and reflectance consistency}

\STATE Compute the L1 loss $\mathcal{L}_{l1}$ for $\mathcal{L}_{l1}(I_l, I_l^{'})$ and $\mathcal{L}_{l1}(I_n, I_n^{'})$
\ENDFOR

\FOR {\textbf{the identity regularization}}

\STATE Input $I_l$ and obtain the generated low-light image $\hat{I}_l$ by:

$\hat{I}_l = D_{d}(U(E_d(I_l),(T_d,I_{l,v})))$

\STATE \textbf{Do caption and reflectance consistency}

\STATE Input $I_n$ and obtain the generated normal-light image $\hat{I}_n$ by:

$\hat{I}_n = D_{l}(U(E_l(I_l),(T_l,I_{n,v})))$

\STATE \textbf{Do caption and reflectance consistency}

\STATE Compute the L1 loss $\mathcal{L}_{l1}$ for $\mathcal{L}_{l1}(I_l, \hat{I}_l)$ and $\mathcal{L}_{l1}(I_n, \hat{I}_n)$
\ENDFOR

\FOR {\textbf{the discriminator learning}}

\STATE learn from the fake predictions ${Dis}_l(\hat{I}_n)$ and ${Dis}_d({\hat{I}_l})$

\STATE learn from the real inputs ${Dis}_d({I_l})$ and ${Dis}_l({I_n})$
\ENDFOR

\ENDFOR
\end{algorithmic}
\end{algorithm}

\clearpage

\begin{algorithm}[ht]
\caption{Caption and Reflectance Consistency}
\label{alg:al_2}
\begin{algorithmic}[1]
\STATE \textbf{Input:}

(1) the caption prompt ${Cap}_l$ and ${Cap}_n$ from $I_l$ and $I_n$, respectively.

(2) the reflectance map $I_{ref,l}$ and $I_{ref,n}$ from $I_l$ and $I_n$, respectively.

\STATE \textbf{Networks:} The reflectance decoder $D_{r}$.

\STATE \textbf{Loss:} The cosine similarity loss $\mathcal{COS}$, L1 loss $\mathcal{L}_{l1}$, and MSE loss $\mathcal{L}_{mse}$.

\FOR{$i$ in $1:iterations$}
\FOR {\textbf{the cycle generation}}

\STATE Input $I_l$ and compute the caption consistency loss $\mathcal{L}_{cap, I_l}$ and reflectance consistency loss $\mathcal{L}_{ref, I_l}$ by:

$Z_l = U(E_l(I_l),(T_l,I_{l,v}^{r}))$

$Z_d = U(E_d(\hat{I}_n),(T_d,I_{l,v}))$

$ \mathcal{L}_{cap, I_l} = \mathcal{COS}(U(E_l(I_l),{Cap}_l), Z_d)$

$\mathcal{L}_{ref, I_l}=\mathcal{L}_{mse}(D_r(Z_l), D_r(Z_d))+\mathcal{L}_{l1}(D_r(Z_d),I_{ref,l})$

\STATE Input $I_n$ and compute the caption consistency loss $\mathcal{L}_{cap, I_n}$ and reflectance consistency loss $\mathcal{L}_{ref, I_n}$ by:

$Z_d = U(E_d(I_n),(T_d,I_{n,v}^{r}))$

$Z_l = U(E_l(\hat{I}_l),(T_l,I_{n,v}))$

$ \mathcal{L}_{cap, I_n} = \mathcal{COS}(U(E_d(I_n),{Cap}_n), Z_l)$

$ \mathcal{L}_{ref, I_n}=\mathcal{L}_{mse}(D_r(Z_d), D_r(Z_l))+\mathcal{L}_{l1}(D_r(Z_l),I_{ref,n})$
\ENDFOR

\FOR {\textbf{the identity regularization}}

\STATE Input $I_l$ and compute the caption consistency loss $\mathcal{L}_{cap, I_l}$ and reflectance consistency loss $\mathcal{L}_{ref, I_l}$ by:

$Z_d = U(E_d(I_l),(T_d,I_{l,v}))$

$ \mathcal{L}_{cap, I_l} = \mathcal{COS}(U(E_l(I_l),{Cap}_l), Z_d)$

$ \mathcal{L}_{ref, I_l}= \mathcal{L}_{l1}(D_r(Z_d),I_{ref,l})$

\STATE Input $I_n$ and compute the caption consistency loss $\mathcal{L}_{cap, I_n}$ and reflectance consistency loss $\mathcal{L}_{ref, I_n}$ by:

$Z_l = U(E_l(I_n),(T_l,I_{n,v}))$

$ \mathcal{L}_{cap, I_n} = \mathcal{COS}(U(E_d(I_n),{Cap}_n), Z_l)$

$ \mathcal{L}_{ref, I_n}= \mathcal{L}_{l1}(D_r(Z_l),I_{ref,n})$

\ENDFOR
\ENDFOR
\end{algorithmic}
\end{algorithm}

% \subsection{Text Prompt}
% For the text prompt, we use ``\textit{normal light photo.}" and ``\textit{low light photo.}" in the paper. During training, we found that providing longer prompts to Stable Diffusion will help improve the performance of the model as shown in \cref{tab:id_1}, so the text prompts we used are:
% \begin{itemize}
% \item natural light, bright lighting, soft illumination, high light, evenly lit, clear visibility, daylight, bright atmosphere, well-lit photo.
% \item low light, dim lighting, low illumination, underexposed, ambient shadows, subdued light, silhouetted, dark atmosphere photo.
% \end{itemize}

% \begin{table}[ht]
%   \centering
%   \caption{Ablation study on the text prompt.}
%   \renewcommand{\arraystretch}{1.2}
%     \scalebox{0.8}{
%     \begin{tabular}{c|ccc|c|c}
%     \toprule
%     \multirow{2}[4]{*}{Prompt} & \multicolumn{3}{c|}{LSRW} & CODaN & DARK FACE \\
% \cmidrule{2-6}          & PSNR $\uparrow$ & SSIM $\uparrow$ & LPIPS $\downarrow$  & Top-1(\%) & mAP(\% \\
%     \midrule
%     Short & 18.16 & 0.5240 & 0.2877 & 59.76 & 16.5 \\
%     Long & 18.41 & 0.5341 & 0.2974 & 60.92 & 16.9 \\
%     \bottomrule
%     \end{tabular}%
%     }
%   \label{tab:id_1}%
% \end{table}%

\clearpage
\section{Experimental Comparisons}

\subsection{Detailed Quantitative Analysis Results.}
Since training datasets used by unsupervised low-light enhancement methods are different, we follow \cite{wang2024zero} to explain training sets of all methods. In the paper, we only show results of RUAS\cite{liu2021retinex} and SCI\cite{ma2022toward} trained on LOL\cite{yang2021sparse}. We can see from \cref{tab:sm_ex_1} that our method performs best on high-level vision tasks and shows the best generalization.
We also show the result of the model trained on the LSRW\cite{hai2023r2rnet} dataset, where we can see that its performance on high-level vision tasks is not as good as that trained on EnlightGan\cite{jiang2021enlightengan}, but still outperforms most existing low-light enhancement methods.

\begin{table}[ht]
  \centering
  \caption{Compare with existing low-light enhancement methods. `T', `S', and `U' indicate traditional, supervised, and unsupervised methods, respectively. $\ast$ denotes our re-implementation with the same training data we use. The best results are highlighted in \textbf{bold}.}
      \renewcommand{\arraystretch}{1.2}
    \resizebox{\linewidth}{!}{
    \begin{tabular}{c|c|c|c|ccc|ccc|c|c|c}
    \toprule
    \multirow{2}[4]{*}{\textbf{Type}} & \multirow{2}[4]{*}{\textbf{Method}} & \multirow{2}[4]{*}{Venue \& Years} & \multirow{2}[4]{*}{Train Set} & \multicolumn{3}{c|}{LSRW\cite{hai2023r2rnet}} & \multicolumn{3}{c|}{LOL\cite{yang2021sparse}} & CODaN\cite{lengyel2021zero} & DARK FACE\cite{poor_visibility_benchmark} & BDD100K-night\cite{yu2020bdd100k} \\
\cmidrule{5-13}          &       &       &       & PSNR $\uparrow$ & SSIM $\uparrow$ & LPIPS $\downarrow$  & PSNR $\uparrow$ & SSIM $\uparrow$ & \multicolumn{1}{c|}{LPIPS $\downarrow$} & Top-1(\%) & mAP(\%) & mIoU(\%) \\
    \midrule
    \multirow{2}[2]{*}{\textbf{T}} & LIME \cite{guo2016lime} & TIP'16 & N/A & 14.88 & 0.3487 & 0.4030 & 16.90 & 0.4917  & 0.4022 & 14.09 & 11.0 & 14.2\\
    & DUAL \cite{zhang2019dual} & CGF'19  & N/A  & 13.76 &  0.3532 & 0.4150 & 16.76 & 0.4911 &  0.4060 & 14.67  & 11.0      &  14.1   \\
    \midrule
    \multirow{2}[2]{*}{\textbf{S}} & RetinexNet \cite{wei2018deepretinexdecompositionlowlight} & BMCV'18 & LOL & 15.59   & 0.4176      & 0.3998  & 17.68   & 0.6477   & 0.4433    & 47.48  & 13.2 & 13.2  \\
          & Retinexformer  \cite{cai2023retinexformer} &ICCV'23   & LOL  & 17.19   & 0.5093   & 0.3314    & 22.79     & 0.8397  & 0.1707 &  52.81 & 16.4 & 15.9  \\
        & CIDNet \cite{yan2025hvi} & CVPR'25 & LOL & 18.00 & 0.5198 & 0.2962 & 20.68 & 0.8411 & 0.1156 & 58.32      & 14.5    & 17.4  \\
    \midrule
    \multirow{16}[2]{*}{\textbf{U}} & EnlightenGan \cite{jiang2021enlightengan} &  TIP'21 & own data   & 17.59 & 0.4867 & 0.3117 & 18.68 & 0.6728 & 0.3013 & 56.42 & 14.2  & 16.6 \\
          & Zero-DCE \cite{guo2020zero} & CVPR'20  & own data  & 15.86 & 0.4536 & 0.3176 & 18.06 & 0.5736 & 0.3125 & 57.76 & 15.9 & 16.6 \\
          & Zero-DCE++ \cite{li2021learning} &  TPAMI'21 & own data  & 16.21 & 0.4571 & 0.3266 & 17.37 & 0.4373 & 0.3118 & 59.88  & 15.2 & 17.7 \\
          & RUAS\_upe \cite{liu2021retinex}  & CVPR'21 & MIT      & 13.00 & 0.3442 & 0.3989  & 13.97 & 0.4656 & 0.3401 & 57.26 & 12.8 & 18.6 \\
          & RUAS\_lol \cite{liu2021retinex}  & CVPR'21 & LOL      & 14.33 & 0.4841 & 0.4800  & 15.33 & 0.4876 & 0.3097 & 51.60 & 14.0 & 15.2 \\
          & RUAS\_dark \cite{liu2021retinex}  & CVPR'21  & DARK FACE  & 14.11 & 0.4183 & 0.3811  & 14.89 & 0.4553 & 0.3722 & 55.42 & 12.0 & 16.5 \\
          & SCI\_easy \cite{ma2022toward}  &  CVPR'22     & MIT      & 11.79 & 0.3174 & 0.4004 & 11.98  & 0.3986 & 0.3543 & 59.76 & 14.0  & 17.5 \\
          & SCI\_medium \cite{ma2022toward}  & CVPR'22      & LOL      & 15.24 & 0.4240 & 0.3218 & 17.30  & 0.5335 & 0.3079 & 58.84 & 14.7  & 18.0 \\
          & SCI\_difficult  \cite{ma2022toward} & CVPR'22      & DARK FACE      & 15.16 & 0.4080 & 0.3259 & 17.25  & 0.5462 & 0.3171 & 59.56 & 14.8  & 17.4 \\
          & PairLlE \cite{fu2023learning}  &  CVPR'23 & own data & 17.60  & 0.5118 & 0.3290 & 19.88 & 0.7777 & 0.2341 & 52.29  & 16.0  & 16.4 \\
          & SADG \cite{zheng2023learning}  & AAAI'23 & own data & 16.32 & 0.4564 & 0.3471 & 16.93 & 0.5372 & 0.3513 & 56.80 & 14.9 & 14.8 \\
          & CLIP-LIT \cite{liang2023iterative} & ICCV'23 & own data      & 13.47 & 0.4089 & 0.3572 & 15.18 & 0.5290 & 0.3689 & 54.64 & 14.1 & 17.3 \\
          & NeRCo \cite{yang2023implicit} & ICCV'23 & LSRW & \textbf{19.46} & 0.5506 & 0.3052 & 19.66 & 0.7172 & 0.2705 & 54.15 & 12.4 & 18.1 \\
          & QuadPrior \cite{wang2024zero} & CVPR'24 & COCO & 16.90  & 0.5429 & 0.3459 & 20.30  & 0.7909  & \textbf{0.1858} & 59.48 & 15.7 & 14.9 \\
          & ZERO-IG\_LSRW \cite{shi2024zero} & CVPR'24 & LSRW & 18.21 & \textbf{0.5665} & 0.4946 & 18.65 & 0.4819 & 0.3819 & 47.60  & 15.6 & 14.9 \\
          & ZERO-IG\_LOL \cite{shi2024zero} & CVPR'24 & LOL & 16.44 & 0.5033 & 0.3744 & 18.13 & 0.7455 & 0.2478 & 53.48 & 15.2 & 14.7 \\
          & LightenDiffusion \cite{jiang2024lightendiffusion} & ECCV'24 & own data & 18.42 & 0.5334 & 0.3209 & 22.79 & 0.8540 & 0.1666 & 57.40  & 16.3 & 16.0 \\
          & LightenDiffusion$^\ast$ & ECCV'24 & EnlightenGan data & 16.92   & 0.5250 & 0.3824 & 18.27  & 0.7944  & 0.2457  & 57.32  & 16.4 & 16.8  \\
          \rowcolor{gray!30}
          & Ours  &       & EnlightenGan data      & 18.41 & 0.5341 & 0.2974 & \textbf{21.32} & \textbf{0.8073} & 0.1928 & \textbf{60.92} & \textbf{16.9} & \textbf{20.1} \\
          \rowcolor{gray!30}
          & Ours-LSRW  &       & LSRW  & 18.96 & 0.5438 & \textbf{0.2673} & 20.22 & 0.7649 & 0.2157 & 60.56 & 16.3 & 18.0\\
    \bottomrule
    \end{tabular}%
  \label{tab:sm_ex_1}%
  }
\end{table}%

\clearpage
\subsection{Visual Quality Comparison.} 
\begin{figure}[ht]
\centering
\includegraphics[width=1\linewidth]{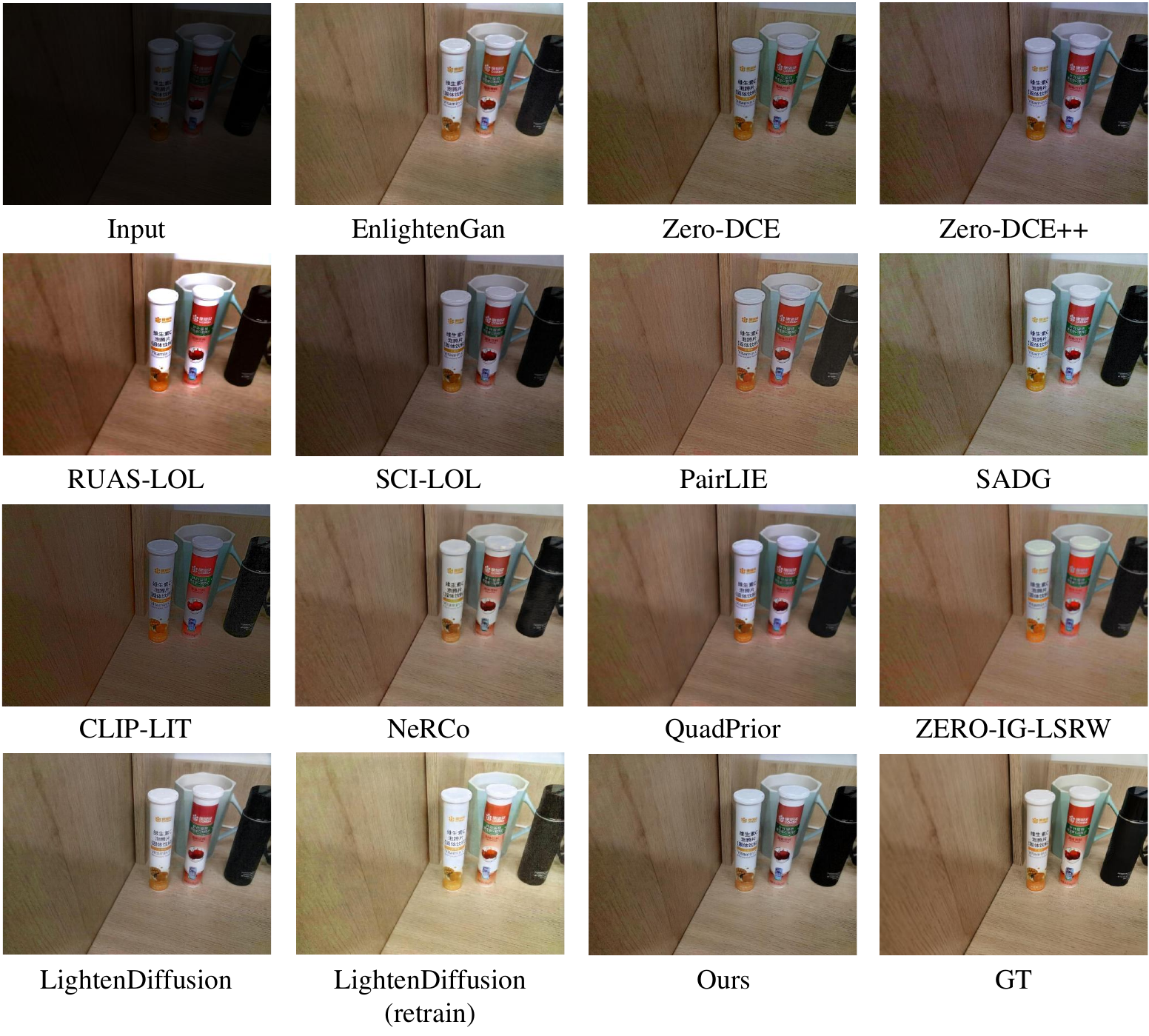}
\caption{Visual quality comparison between the proposed method and other state-of-the-art methods on the LSRW\cite{hai2023r2rnet}.}
\label{fig:sm_ex_1}
% \vspace{-1em}
\end{figure}

\begin{figure}[ht]
\centering
\includegraphics[width=1\linewidth]{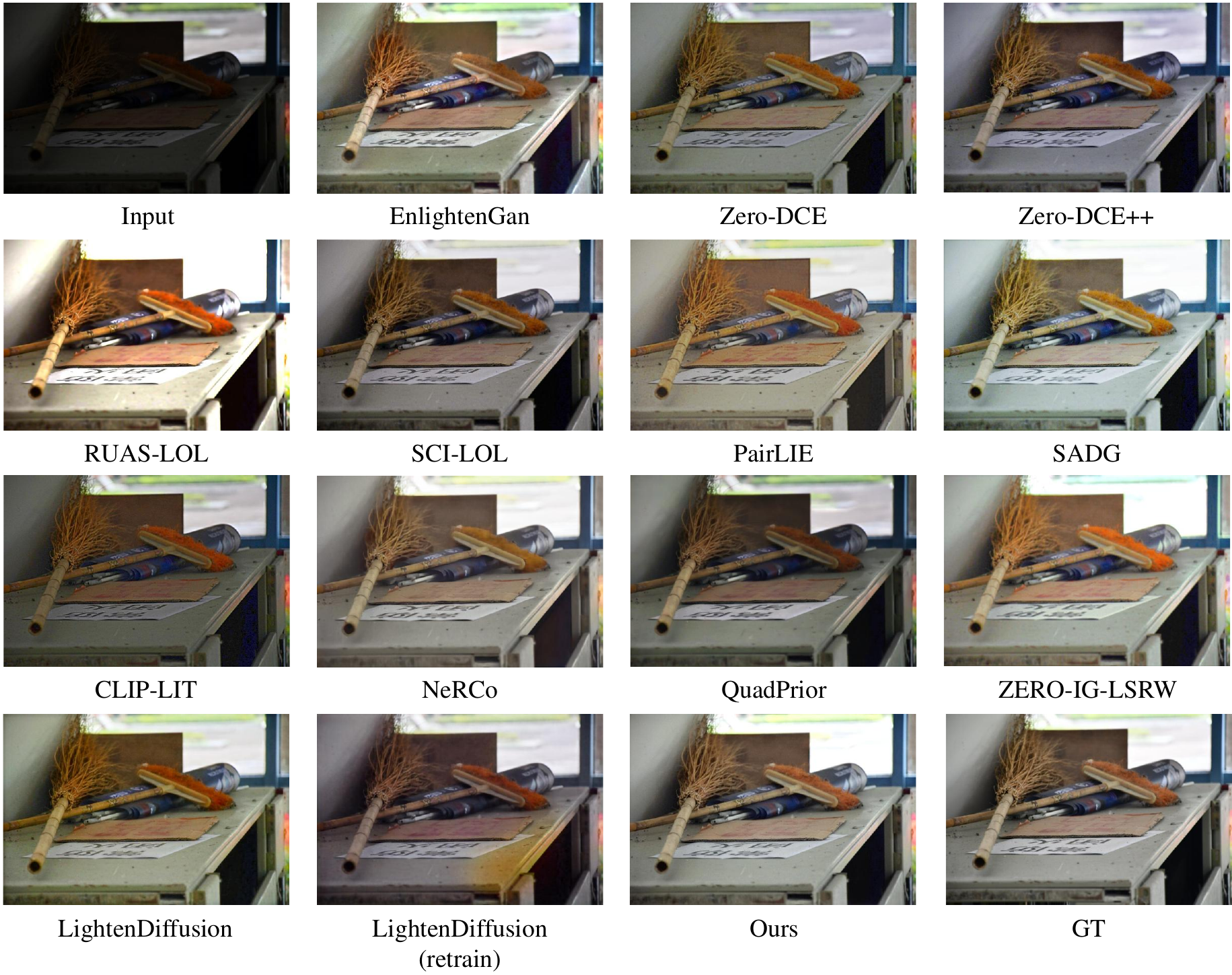}
\caption{Visual quality comparison between the proposed method and other state-of-the-art methods on the LSRW\cite{hai2023r2rnet}.}
\label{fig:sm_ex_2}
% \vspace{-1em}
\end{figure}

\begin{figure}[ht]
\centering
\includegraphics[width=1\linewidth]{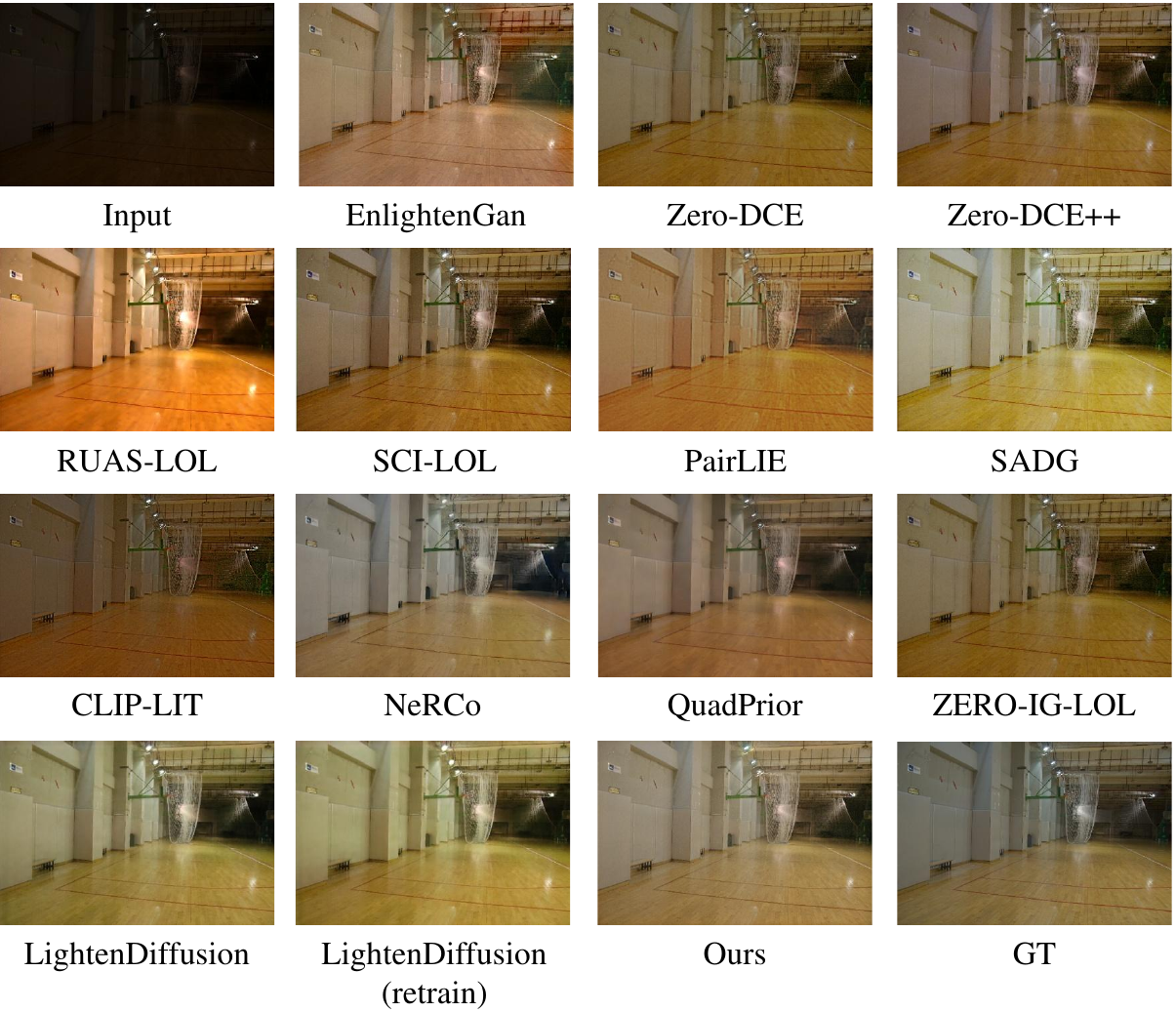}
\caption{Visual quality comparison between the proposed method and other state-of-the-art methods on the LOL\cite{yang2021sparse}.}
\label{fig:sm_ex_3}
% \vspace{-1em}
\end{figure}

We also show enhancement results of different low-light enhancement methods, as shown in \cref{fig:sm_ex_1}, \cref{fig:sm_ex_2} and \cref{fig:sm_ex_3}.
% We find that although NeRCo\cite{yang2023implicit} achieve the best performance on LSRW in quantitative analysis, with a PSNR of 19.46, its enhanced images had obvious distortion, such as the text on the bottle in \cref{fig:sm_ex_1} and the bristles of the broom in \cref{fig:sm_ex_2}. 
% Similarly, LightenDiffusion\cite{jiang2024lightendiffusion} achieves the best performance on LOL, but its results in \cref{fig:sm_ex_3} are quite different from GT. The performance of its retrained model does not improve. 
Our model achieves relatively high-fidelity results. 
% This also shows that image quality evaluation indicators such as PSNR and SSIM cannot perfectly evaluate the quality of the generated images.

\clearpage
\subsection{High-level Vision Comparison.}
\begin{figure}[ht]
\centering
\includegraphics[width=0.8\linewidth]{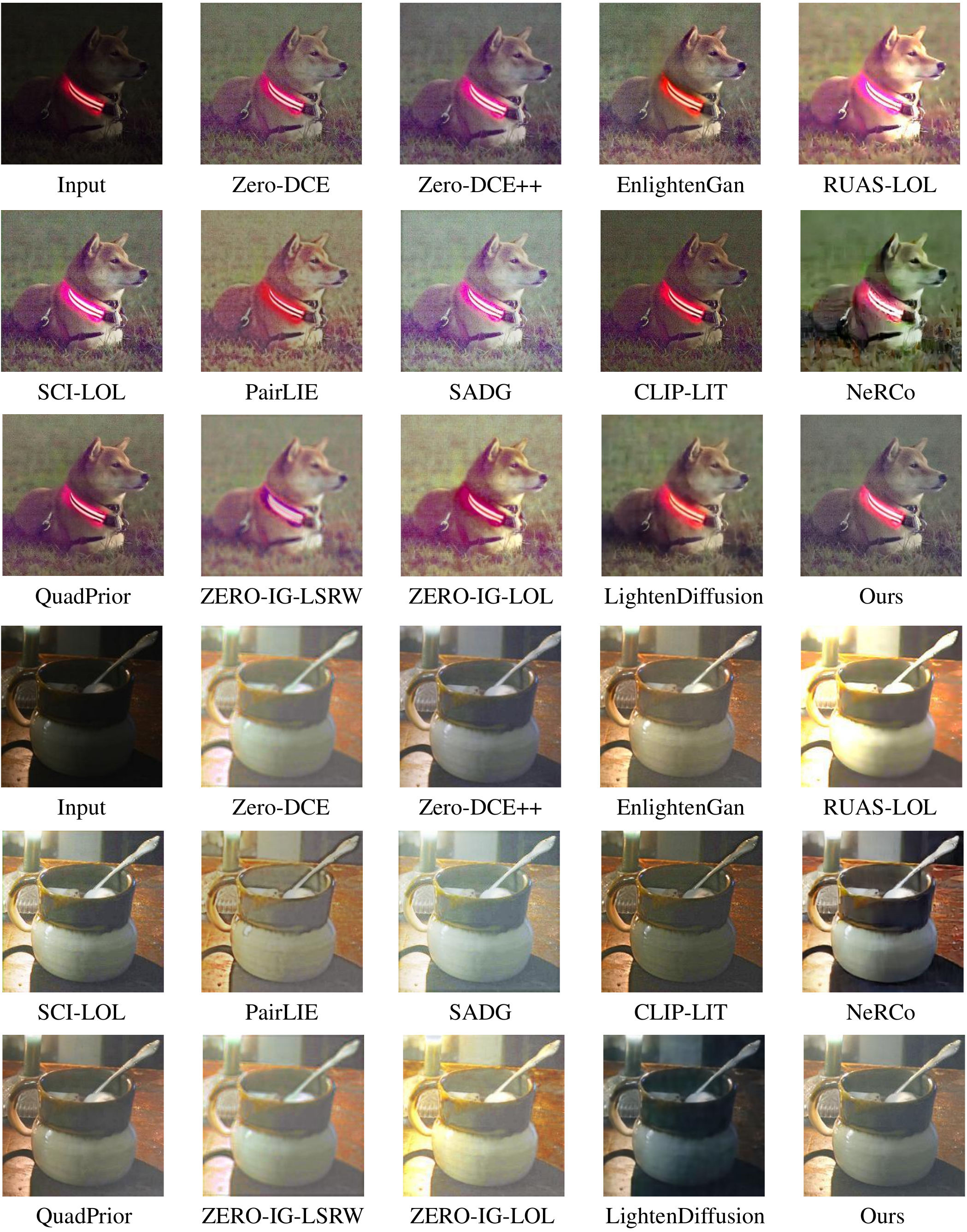}
\caption{Qualitative comparison of the proposed method with other state-of-the-art low-light enhancement methods on night image classification on CODaN\cite{lengyel2021zero}.}
\label{fig:sm_ex_7}
% \vspace{-1em}
\end{figure}

\begin{figure}[ht]
\centering
\includegraphics[width=1\linewidth]{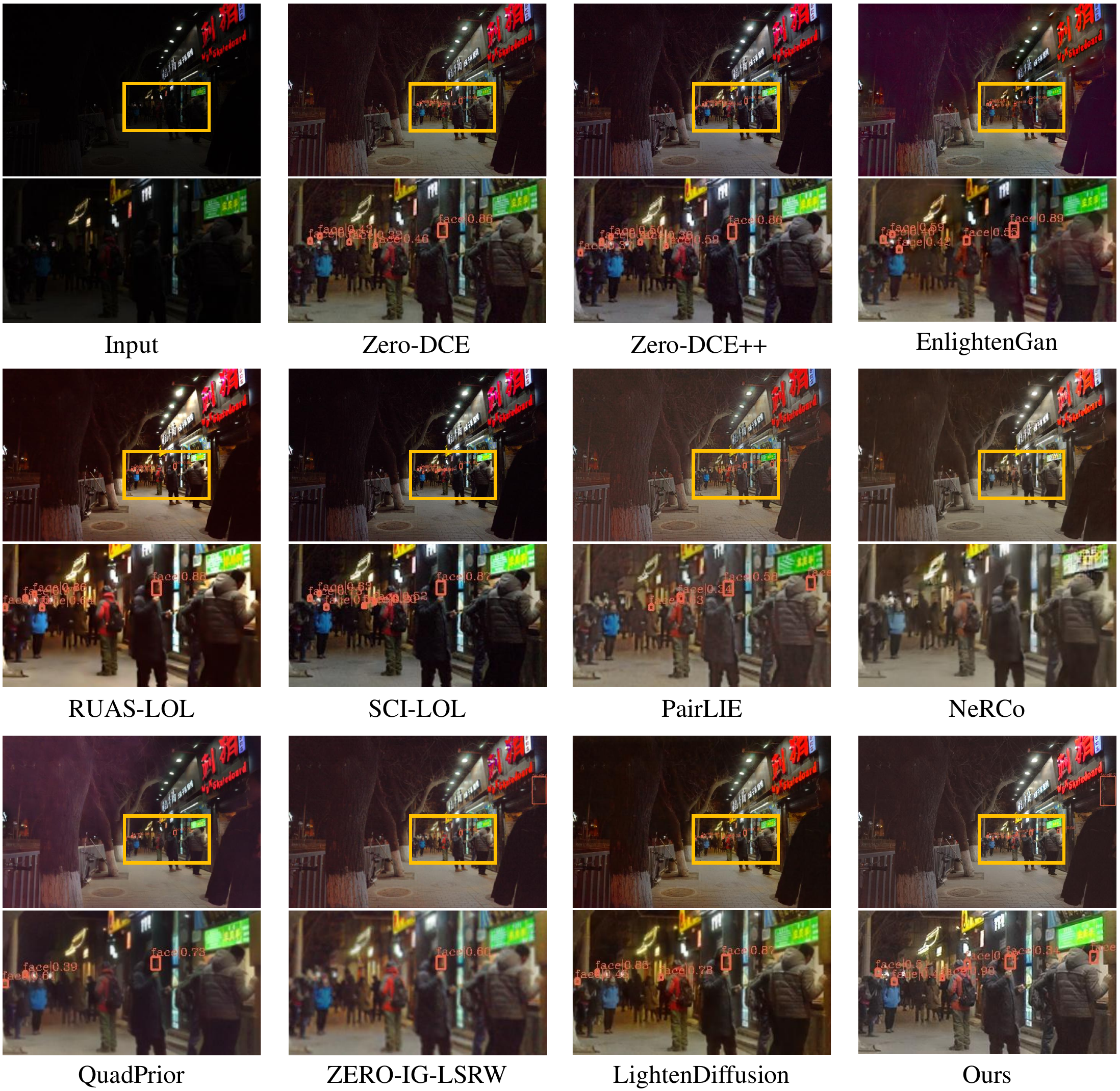}
\caption{Qualitative comparison of the proposed method with other state-of-the-art low-light enhancement methods on dark face detection on DARK FACE\cite{poor_visibility_benchmark}.}
\label{fig:sm_ex_4}
% \vspace{-1em}
\end{figure}

\begin{figure}[ht]
\centering
\includegraphics[width=1\linewidth]{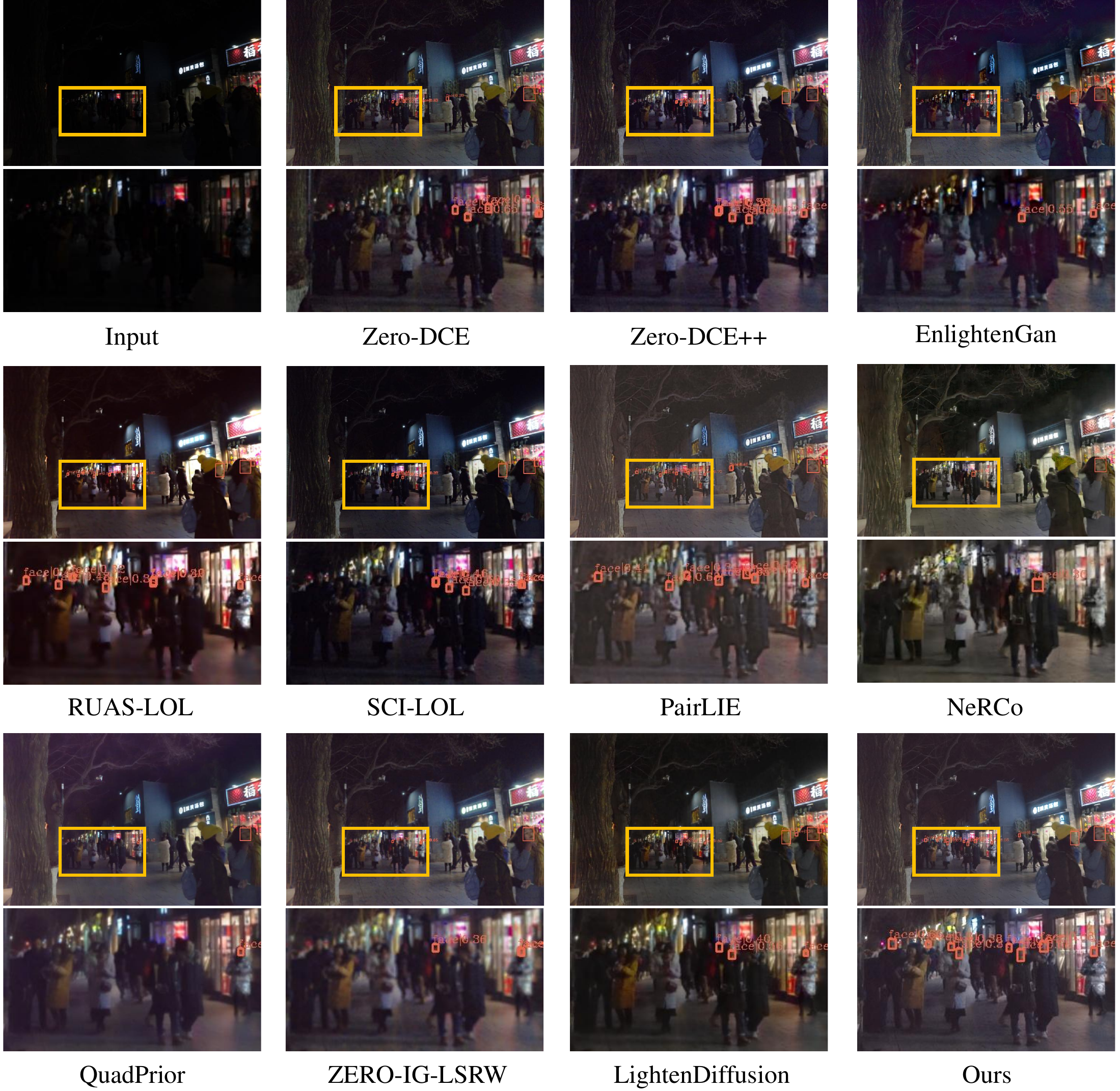}
\caption{Qualitative comparison of the proposed method with other state-of-the-art low-light enhancement methods on dark face detection on DARK FACE\cite{poor_visibility_benchmark}.}
\label{fig:sm_ex_5}
% \vspace{-1em}
\end{figure}

\begin{figure}[ht]
\centering
\includegraphics[width=1\linewidth]{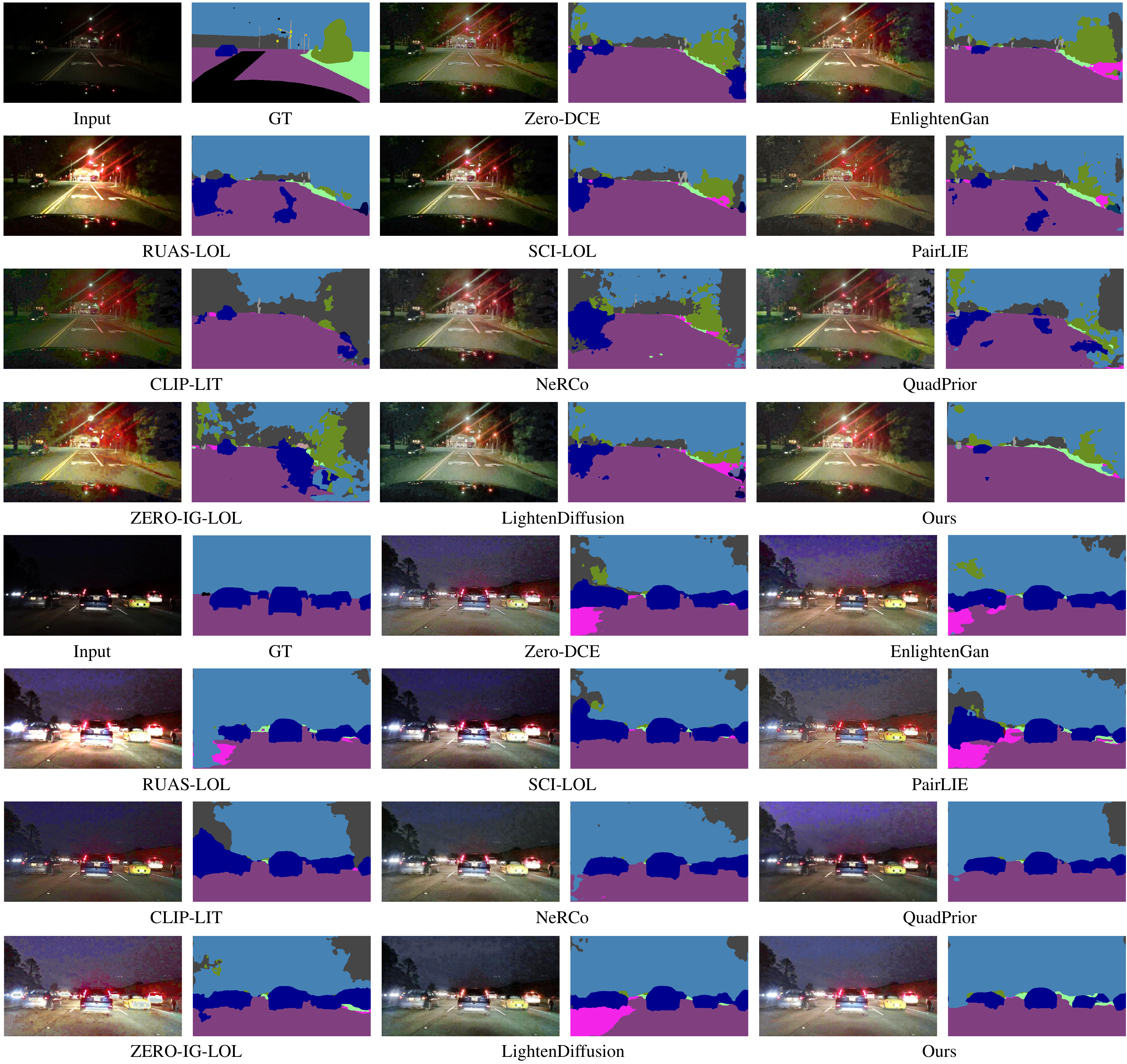}
\caption{Qualitative comparison of the proposed method with other state-of-the-art low-light enhancement methods on nighttime image semantic segmentation on BDD100k-night\cite{yu2020bdd100k}.}
\label{fig:sm_ex_6}
% \vspace{-1em}
\end{figure}

We show results of existing low-light enhancement methods on night image classification in \cref{fig:sm_ex_7}, low-light face detection in \cref{fig:sm_ex_4} and \cref{fig:sm_ex_5}, and nighttime semantic segmentation in \cref{fig:sm_ex_6}. We can see that our method achieves the best performance.

\clearpage
\subsection{Visual Analysis of Different Adapters.}
\begin{figure}[ht]
\centering
\includegraphics[width=0.6\linewidth]{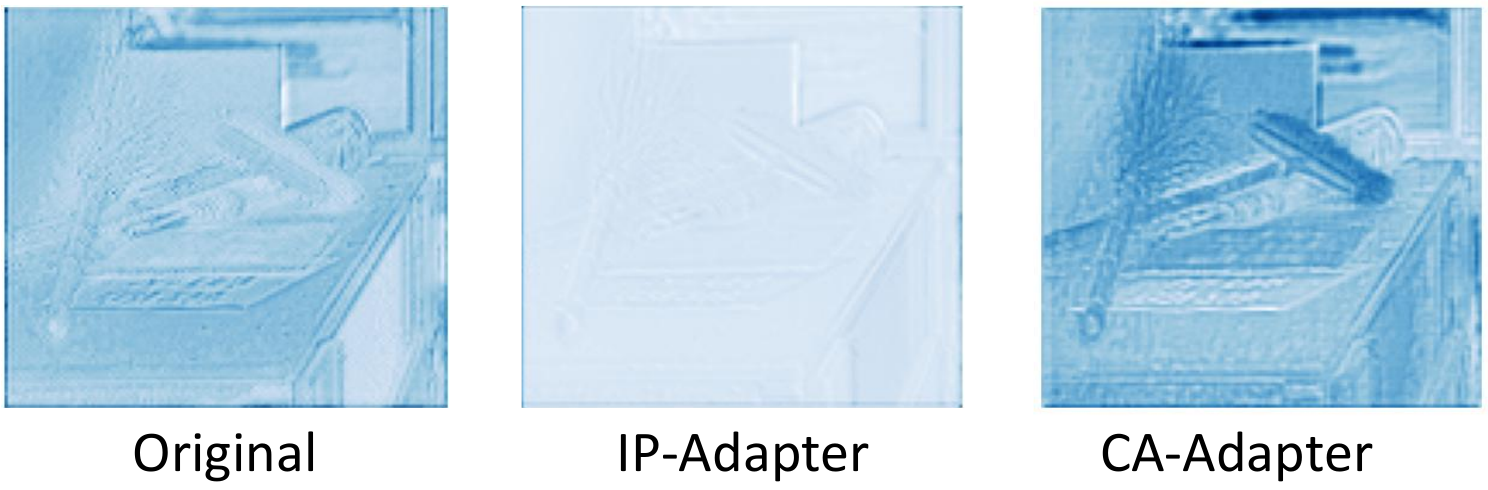}
\caption{Visual analysis of the different adapters.}
\label{fig:sm_ex_8}
% \vspace{-1em}
\end{figure}

In the paper, we compared the visual analysis of image feature extraction by different components. We supplement the visual analysis experiments using different adapters. 
As shown in \cref{fig:sm_ex_8}, results using the IP-Adapter\cite{ye2023ip} are even worse than the original adapter that directly connects text and image features. 
This fully demonstrates that the IP-Adapter is not suitable for the illumination-aware image prompt. The cycle-attention adapter we proposed fully exploits the semantic features of the illumination-aware image prompt and achieves the best result.

\subsection{Another Baseline on Normal Light Images.}
The pre-trained downstream models tend to overfit on training data, such as classification and detection results as shown in \cref{tab:b5}, while the normal light segmentation results are close to our enhanced low light images because they are tested on BDD100k, which the model has not seen.

\begin{table}[h]
  \centering
  % \caption{Baseline comparison.}
  % \vspace{-0.8em}
  \begin{tabular}{c|c|c|c}
    \hline
    Setting & Cls Top-1(\%) & Det mAP(\%) & Seg mIoU(\%) \\
    \hline
    Pretrained Data & CODaN-day & WIDER FACE & Cityscapes \\
    \hline
    Normal light Data & CODaN-day & WIDER FACE & BDD100k-day \\
    Baseline (Normal) & 82.52 & 55.9 & 23.1 \\
    \hline
    Low light Data & CODaN-dark & DARK FACE & BDD100k-night \\
    Baseline (Low) & 53.24 & 10.8 & 11.4 \\
    Ours & 60.92 & 16.9 & 20.1 \\
    \hline
  \end{tabular}
  \label{tab:b5}
\end{table}

\section{Failure cases.}
\begin{figure}[h]
    \centering
    \includegraphics[width=0.8\linewidth]{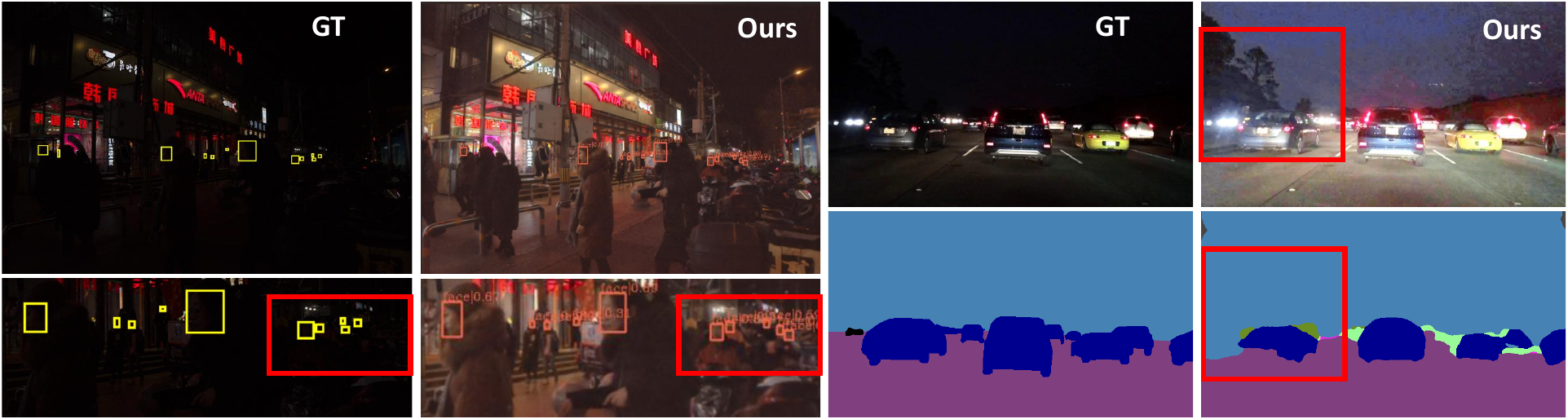}
  \caption{Fail cases.}
    \label{fig:c1}%
\end{figure}

While our method generalizes better than existing approaches, two challenges remain as shown in \cref{fig:c1}. First, due to detail loss in low-light images, small object detection remains difficult, which is a limitation across most methods. Second, under extreme low-light degradation, enhanced outputs may still contain noise and artifacts, hindering complete restoration (e.g., the tree in segmentation). Transferring semantic knowledge from normal-light conditions to guide restoration remains a challenging problem.
Future work needs to explore more diverse low-light scenarios and the performance upper bound.

\end{document}